\documentclass[]{template}

\usepackage[utf8]{inputenc}             
\usepackage[T1]{fontenc}                
\usepackage{url}                        
\usepackage{booktabs}                   
\usepackage{multirow}
\usepackage{colortbl}
\usepackage{multicol}
\usepackage{amsfonts}                   
\usepackage{nicefrac}                   
\usepackage{microtype}                  
\usepackage[dvipsnames]{xcolor}         

\usepackage{latexsym}

\usepackage{graphicx}
\usepackage{float}
\usepackage{subcaption}
\usepackage{wrapfig}
\usepackage{lipsum}

\usepackage{bm}

\usepackage{tabularx} 
\usepackage{ragged2e} 
\newcolumntype{L}{>{\RaggedRight\hangafter=1\hangindent=0em}X}

\usepackage{enumitem}

\usepackage{amsmath}
\usepackage{amssymb}
\usepackage{mathtools}
\usepackage{amsthm}

\setboolean{logo}{true}    

\usepackage[linesnumbered,ruled,vlined]{algorithm2e}

\hypersetup{
    colorlinks=true,
    linkcolor=red,
    citecolor=Cerulean,
    filecolor=magenta,      
    urlcolor=magenta,
}

\usepackage[capitalize,noabbrev]{cleveref}
\crefname{section}{§}{§§}
\Crefname{section}{§}{§§}

\usepackage{calligra}
\DeclareMathAlphabet{\mathcalligra}{T1}{calligra}{m}{n}

\usepackage{pifont}

\theoremstyle{plain}

\theoremstyle{definition}

\theoremstyle{remark}

\renewcommand{\paragraph}[1]{\vspace{1mm}\noindent\textbf{#1}}

\DeclareCaptionLabelFormat{cont}{#1~#2\alph{ContinuedFloat}}
\usepackage[most]{tcolorbox}
\tcbset{
  promptbox/.style={
    top=10pt,
    colback=lightgray!20,
    colframe=Black,
    colbacktitle=NavyBlue,
    enhanced,
    center,
    attach boxed title to top center={yshift=-0.1in,xshift=0.0in},
    boxed title style={boxrule=0pt,colframe=white,},
  }
}
\newtcolorbox{promptbox}[2][]{promptbox, title=#2,#1}
\tcbset{
  takeawaybox/.style={
    top=10pt,
    colback=lightgray!20,
    colframe=Black,
    colbacktitle=BurntOrange,
    enhanced,
    center,
    attach boxed title to top center={yshift=-0.1in,xshift=0.0in},
    boxed title style={boxrule=0pt,colframe=white,},
  }
}
\newtcolorbox{takeawaybox}[2][]{takeawaybox, title=#2,#1}
\tcbset{
  observationbox/.style={
    top=10pt,
    colback=lightgray!20,
    colframe=Black,
    colbacktitle=YellowGreen,
    enhanced,
    center,
    attach boxed title to top center={yshift=-0.1in,xshift=0.0in},
    boxed title style={boxrule=0pt,colframe=white,},
  }
}
\newtcolorbox{observationbox}[2][]{observationbox, title=#2,#1}

\usepackage{xspace}

\newcommand\blfootnote[1]{%
  \begingroup
  \renewcommand\thefootnote{}\footnote{#1}%
  \addtocounter{footnote}{-1}%
  \endgroup
}

\usepackage{CJK}

\usepackage{xcolor}
\usepackage{hyperref}
\usepackage{url}
\usepackage{graphicx}
\usepackage{xspace}
\usepackage{colortbl}
\usepackage{booktabs}
\usepackage{multirow,marvosym}
\usepackage{wrapfig}

\definecolor{ForestGreen}{rgb}{0, 0.69, 0.31}
\definecolor{NavyBlue}{rgb}{0, 0.44, 0.75}
\newcommand{\hgreen}[1]{\textcolor{ForestGreen}{\textbf{#1}}} 

\usepackage{pifont}

\definecolor{00blue}{RGB}{139,169,235}
\newcommand{\methodname}{\textcolor{00blue}{Visual-ERM}\xspace}
\newcommand{\benchname}{VC-RewardBench\xspace}

\definecolor{Cerulean}{rgb}{0.0, 0.48, 0.65}

\definecolor{customblue}{HTML}{005AD7}

\title{Visual-ERM: Reward Modeling for Visual Equivalence}

\author[1,2]{Ziyu Liu*}
\author[2,3]{Shengyuan Ding*}
\author[2]{Xinyu Fang}
\author[2]{Xuanlang Dai}
\author[2]{Penghui Yang}
\author[2]{Jianze Liang}
\author[2]{Jiaqi Wang}
\author[2]{Kai Chen}
\author[2,4]{Dahua Lin}
\author[2]{Yuhang Zang$\dagger$}

\affil[1]{Shanghai Jiao Tong University} \affil[2]{Shanghai AI Laboratory} \affil[3]{Fudan University} \affil[4]{CUHK}

\begin{abstract}
Vision-to-code tasks require models to reconstruct structured visual inputs, such as charts, tables, and SVGs, into executable or structured representations with high visual fidelity.
While recent Large Vision Language Models (LVLMs) achieve strong results via supervised fine-tuning, reinforcement learning remains challenging due to misaligned reward signals.
Existing rewards either rely on textual rules or coarse visual embedding similarity, both of which fail to capture fine-grained visual discrepancies and are vulnerable to reward hacking.
We propose the Visual Equivalence Reward Model (\methodname), a multimodal generative reward model that provides fine-grained, interpretable, and task-agnostic feedback to evaluate vision-to-code quality directly in the rendered visual space.
Integrated into Reinforcement Learning (RL), \methodname improves Qwen3-VL-8B-Instruct by +8.4 on chart-to-code and yields consistent gains on table and SVG parsing (+2.7, +4.1 on average), and further strengthens test-time scaling via reflection and revision.
We also introduce VisualCritic-RewardBench (\benchname), a benchmark for judging fine-grained image-to-image discrepancies on structured visual data, where \methodname at 8B decisively outperforms Qwen3-VL-235B-Instruct and approaches leading closed-source models.
Our results suggest that fine-grained visual reward supervision is both necessary and sufficient for vision-to-code RL, regardless of task specificity.
\end{abstract}

\begin{document}

\blfootnote{$\dagger$ Corresponding authors: Yuhang Zang (zangyuhang@pjlab.org.cn)}
\blfootnote{$*$ Equal contribution. Code is at \url{https://github.com/InternLM/Visual-ERM}}

\maketitle

\section{Introduction}
\begin{figure*}[t]
    \begin{center}
    \includegraphics[width=1.\linewidth]{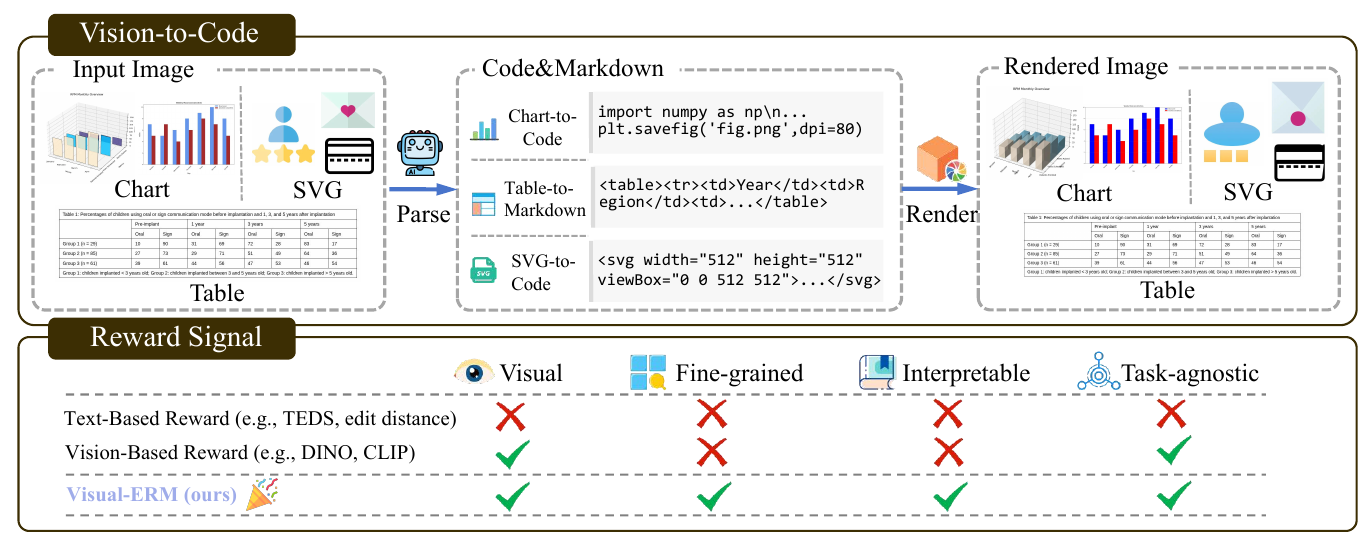}
    \end{center}
    \vspace{-12pt}
    \caption{\small \textbf{Overview of vision-to-code reward modeling and \methodname.}
    \textbf{Top:} vision-to-code parses a structured visual input (chart, table, or SVG) into code or markup, which is rendered back to an image so that quality can be judged in the visual space.
    \textbf{Bottom:} we compare reward signals along four properties: \emph{visual}, \emph{fine-grained}, \emph{interpretable}, and \emph{task-agnostic}. Text-based rewards (e.g., TEDS, edit distance) ignore visual cues; vision-encoder similarities (e.g., DINO~\cite{simeoni2025dinov3}, CLIP~\cite{radford2021learning}) capture visual content but remain coarse and opaque. \methodname combines all four properties, providing a reliable supervisor across vision-to-code tasks.}
    \label{fig:teaser}
    \vspace{-12pt}
\end{figure*}

Large Vision Language Models (LVLMs) have made rapid progress in multimodal understanding \cite{OpenAI_O1,OpenAI_O3,Qwen3-VL}.
Among the resulting capabilities, \textbf{vision-to-code} (Fig.~\ref{fig:teaser} top) stands out as a particularly important primitive: it converts structured visual inputs such as charts, tables, and SVGs into executable code \cite{zhao2025vincicoder} or markdown \cite{niu2025mineru2}.
Vision-to-code has become a key primitive for downstream uses including UI-to-code generation \cite{si2025design2code}, scientific document parsing, and knowledge management.

Most existing work improves vision-to-code with supervised fine-tuning (SFT) \cite{zhao2025vincicoder,zhao2025chartcoder}, which is data-intensive and often \textit{generalizes poorly} across tasks.
Reinforcement learning (RL) is a natural alternative \cite{ling2025table2latex_doc_rl,tan2025chartmaster}, but its effectiveness depends on the \textbf{reward signal}.
A reliable reward for vision-to-code should be \emph{visual} (judge the rendered output rather than the source code), \emph{fine-grained} (penalize element-level errors in layout, axes, and content), \emph{interpretable} (localize what is wrong rather than emit an opaque scalar), and \emph{task-agnostic} (cover charts, tables, and SVGs within a single model).
As shown at the bottom of Fig.~\ref{fig:teaser}, \textbf{current rewards fall short} on these axes.
Text-based rules such as edit distance and Tree-Edit-Distance Similarity (TEDS) operate on predicted code or markup, so they are not \emph{visual}: layout, alignment, and other rendering errors are invisible to them.
Vision-encoder similarities such as DINO~\cite{simeoni2025dinov3} are \emph{visual} but not \emph{fine-grained} or \emph{interpretable}: their encoders are pre-trained for semantic invariance and tolerate small spatial or structural shifts that change the meaning of a chart or table, while collapsing evidence into a single scalar (Sec.~\ref{sec:analysis}).
Neither family is \emph{task-agnostic} across chart, table, and SVG within a single model, leaving a \textbf{gap} between the reward and the underlying notion of \textit{visual equivalence}.
We present the detailed analysis in Sec.~\ref{sec:analysis}.

We address this gap with the Visual Equivalence Reward Model (\methodname), a multimodal generative reward model that scores vision-to-code outputs by \textit{reasoning over the rendered image directly}.
We train \methodname on a discrepancy-annotated corpus of reference--prediction image pairs assembled from two complementary sources: targeted edits that inject pre-defined error types and natural inferences from weaker LVLMs, with fine-grained annotations distilled from a stronger proprietary model.
At deployment, the same model serves two roles: it converts per-error severities into a bounded scalar that, paired with a render-success term, supplies the reward for GRPO-based RL, and it returns the structured discrepancy list as natural-language feedback for test-time scaling.

We evaluate \methodname along three axes.
On RL, we integrate it into vision-to-code pipelines and observe that \methodname boosts Qwen3-VL-8B-Instruct~\cite{Qwen3-VL} by $+8.4$ points on chart-to-code and yields consistent gains on table-to-markdown ($+2.7$) and SVG-to-code ($+4.1$).
To assess the reward signal itself, we introduce VisualCritic-RewardBench (\benchname), a benchmark for fine-grained image-to-image discrepancy judgment on structured visuals; \benchname exposes consistent weaknesses of current LVLMs, and \methodname at 8B \textit{improves over Qwen3-VL-235B-Instruct}, narrowing the gap to leading closed-source models.
For test-time scaling, \methodname provides interpretable feedback that drives reflection and revision, yielding further gains.

Our contributions are: \textbf{1)} We analyze existing reward paradigms for vision-to-code and identify four properties a reliable reward must satisfy, namely \emph{visual}, \emph{fine-grained}, \emph{interpretable}, and \emph{task-agnostic}; we further show that existing rewards each violate at least two of these properties and induce reward hacking under RL. \textbf{2)} Guided by this analysis, we propose \methodname, a multimodal generative reward model that scores rendered outputs with fine-grained, interpretable, and task-agnostic signals, and we integrate it into both RL and test-time scaling. \textbf{3)} We release VisualCritic-RewardBench, a benchmark of $1{,}335$ instances for fine-grained image-to-image discrepancy judgment. \textbf{4)}  \methodname improves Qwen3-VL-8B-Instruct by $+8.4$ on chart-to-code, $+2.7$ on table-to-markdown, and $+4.1$ on SVG-to-code, with further gains from reflection-and-revision at inference time.
\section{Analysis of Reward Signals for Vision-to-Code}\label{sec:analysis}
\begin{figure*}[t]
    \begin{center}
    \includegraphics[width=1.\linewidth]{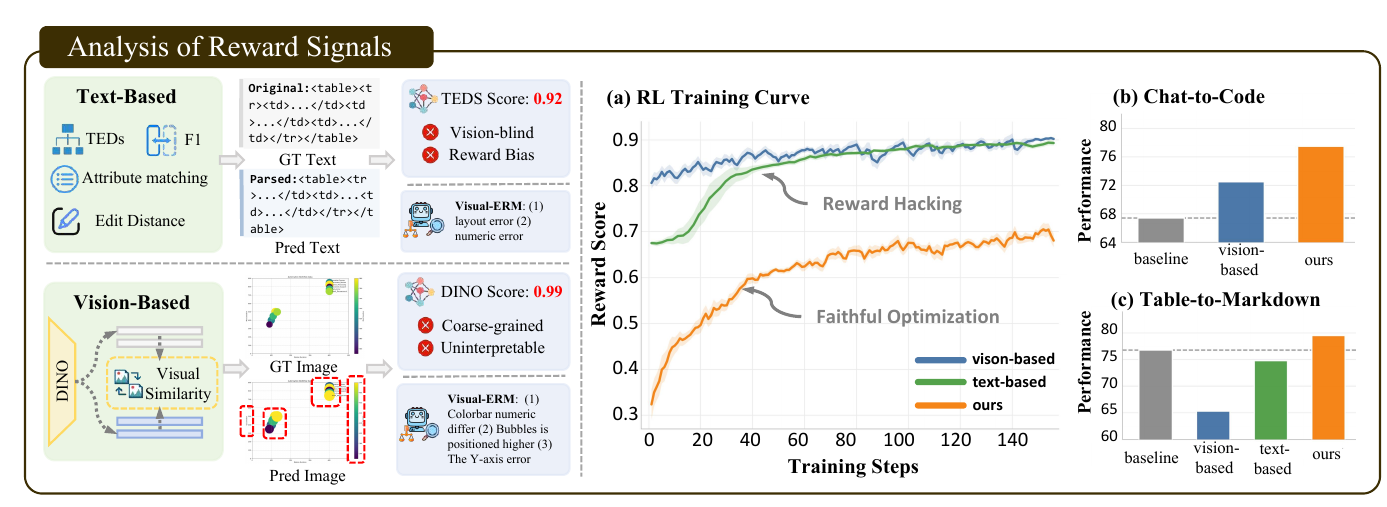}
    \end{center}
    \vspace{-12pt}
    \caption{\small \textbf{Analysis of existing rewards and downstream impact.}
    \textit{Left:} a text-based reward (TEDS) and a vision-based reward (DINO) assign near-perfect scores to predictions that \methodname identifies as visually wrong.
    \textit{Right:} (a) RL training curves where text- and vision-based rewards plateau via reward hacking while \methodname keeps improving; (b, c) downstream gains on Chart-to-Code and Table-to-Markdown.}
    \label{fig:analysis}
    \vspace{-12pt}
\end{figure*}
RL is the most promising route to scale vision-to-code beyond imitation, but RL is only as reliable as its reward.
We first present the limitations of existing reward families in Fig.~\ref{fig:analysis}.

\noindent \textbf{Setup.}
Let $x$ denote the rendered reference image and $\hat{y}_\text{img}$ the rendered prediction.
We define \textit{visual equivalence} as the property that $\hat{y}_\text{img}$ matches $x$ at every level a downstream user would inspect, including layout, axes, labels, and numeric content.
We probe each reward in two complementary ways: qualitative case studies on held-out chart and table examples, and RL training under each reward followed by downstream evaluation.

\noindent \textbf{Text-based rules are blind to rendering.}
TEDS, edit distance, and related metrics compare predicted code or markup tokens against a reference sequence and never inspect $x$ or $\hat{y}_\text{img}$.
A prediction that is textually close but renders to the wrong figure therefore receives a near-perfect score, and a prediction that renders correctly via different yet equivalent code is penalized.
The top-left case in Fig.~\ref{fig:analysis} shows a table-to-markdown output with \textbf{TEDS$\,=\,0.92$} whose rendered version exhibits both header-hierarchy and numeric-cell errors, all of which \methodname localizes.

\noindent \textbf{Vision-encoder similarities are tuned for the wrong invariance and are uninterpretable.}
Vision foundation models such as DINO~\cite{caron2021emerging} and CLIP~\cite{radford2021learning} are pre-trained to be invariant to translation, scale, and other transformations that preserve semantic content.
This invariance is at odds with vision-to-code, where a small spatial shift or a permuted bar order changes the underlying data, and the resulting scalar also hides \emph{which} element is wrong.
The bottom-left case in Fig.~\ref{fig:analysis} shows a chart with \textbf{DINO similarity above $0.99$} for which \methodname enumerates colorbar, bubble-position, and y-axis errors.

\noindent \textbf{Reward hacking under RL.} We optimize a chart-to-code policy against each reward and report the proxy reward score over training in Fig.~\ref{fig:analysis}(a). The text-based and vision-based curves rise quickly and saturate, consistent with reward hacking: the policy maximizes the proxy without improving the rendered output. Optimizing against \methodname keeps improving throughout training, suggesting that its supervision remains aligned with rendered-image fidelity. The next paragraph translates these proxy trends into rendered-task performance.

\noindent \textbf{Downstream impact.} Fig.~\ref{fig:analysis}(b, c) evaluates the resulting policies on Chart-to-Code and Table-to-Markdown. Policies trained with text-based or vision-based rewards do not consistently improve over the SFT baseline, and underperform it on Table-to-Markdown, indicating that the proxy gains in Fig.~\ref{fig:analysis}(a) are illusory. Optimizing against \methodname yields the largest gains in our comparison on both tasks.

\noindent \textbf{Summary.}
Together, the failure cases and the RL probes point to four properties a vision-to-code reward must satisfy:
(i) be visual, operating in the rendered image space, since text-only rewards are blind to rendering;
(ii) be fine-grained and locally sensitive, since coarse global similarity hides element-level errors;
(iii) be interpretable, since a single scalar invites reward hacking and gives the policy no signal about \emph{what} to fix; and
(iv) be task-agnostic, since the same failures recur across charts, tables, and SVGs.
Sec.~\ref{sec:method} instantiates these four properties in \methodname.
\section{Methods}\label{sec:method}
The analysis in Sec.~\ref{sec:analysis} defines four properties for a vision-to-code reward: it must operate in the rendered image space, be fine-grained and locally sensitive, expose interpretable evidence, and remain task-agnostic.
\methodname instantiates these properties through three coupled components, each visualized in Fig.~\ref{fig:framework}: (i) a discrepancy-annotated dataset $\mathcal{D}_{\text{reward}}$ and a single LVLM trained on it; (ii) a deployment-time pipeline that converts \methodname outputs into a scalar reward for RL; and (iii) the same outputs reused as natural-language feedback for test-time scaling.
We also release \benchname, a diagnostic benchmark for fine-grained image-to-image discrepancy judgment.

\noindent \textbf{Notation.}
We retain the symbols introduced in Sec.~\ref{sec:analysis}.
Let $m\!\in\!\{\textsc{Chart},\textsc{Table},\textsc{SVG}\}$ index the task and $\mathcal{R}_m(\cdot)$ its renderer.
A vision-to-code policy $\pi_{\theta}(y\mid x)$ takes the reference image $x$ as input and emits structured text $y$, which is rendered to $\hat{y}_{\text{img}}=\mathcal{R}_m(y)$.
We write $f_{\theta_{\text{ERM}}}$ for \methodname; $y^{\star}$ denotes ground-truth code or markup.

\subsection{Visual-ERM: Data, Annotation, and Training}
\begin{figure*}[t]
    \begin{center}
    \includegraphics[width=1.\linewidth]{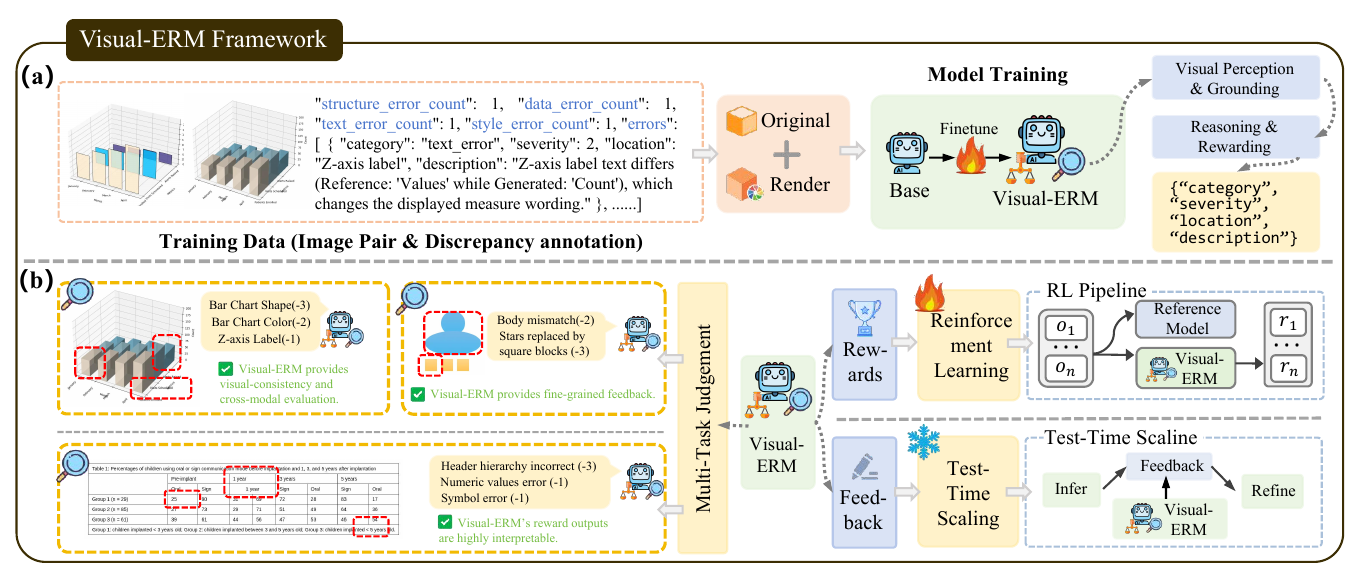}
    \end{center}
    \vspace{-12pt}
    \caption{\small \textbf{(a) Training data and model training.} From paired ground-truth and rendered images, we obtain fine-grained discrepancy annotations (category, severity, location, description) and fine-tune a base LVLM into \methodname.
    \textbf{(b) Deployment.} \methodname produces unified judgments across chart, SVG, and table tasks, yielding scalar rewards for RL and natural-language feedback for test-time scaling.}
    \label{fig:framework}
    \vspace{-12pt}
\end{figure*}

The training pipeline shown in Fig.~\ref{fig:framework}(a) consists of three stages: image-pair construction, fine-grained discrepancy annotation, and supervised fine-tuning.

\noindent \textbf{Image-pair construction.}
We curate pairs $(x,\hat{y}_{\text{img}})$ that span the discrepancies a downstream policy is likely to produce, using two complementary sources.
(1) \emph{Targeted Edit}: high-capacity LVLMs perturb $y^{\star}$ to inject a pre-defined taxonomy of error types, providing systematic coverage of structural, data, textual, and stylistic failures.
(2) \emph{Natural Inference}: weaker LVLMs predict $y$ directly, so the resulting errors match the distribution that an RL policy actually emits during rollout.
Each candidate $y$ is rendered with $\mathcal{R}_m$ to obtain $\hat{y}_{\text{img}}$ and paired with the reference image $x$.

\noindent \textbf{Fine-grained discrepancy annotation.}
For every pair we collect a structured annotation 
$a=\{(\text{category},\text{severity},\\ \text{location},\text{description})_k\}_{k=1}^{K}$ that localizes each discrepancy and grades its impact.
A natural concern is whether large open-source LVLMs already produce such annotations directly; we find they do not.
Even Qwen3-VL-235B-Instruct~\cite{Qwen3-VL} misses subtle structural and textual deviations, as quantified in Sec.~\ref{sec:visual_erm_bench_results}.
We therefore use a stronger model, GPT-5-mini~\cite{singh2025openai}, as a \emph{bootstrap teacher} that proposes candidate discrepancies, which are then filtered by rendering-consistency checks (proposals referring to regions absent in $\hat{y}_{\text{img}}$ are discarded) and by agreement with the ground-truth code $y^{\star}$ at the structural level.
The resulting reward dataset is $\mathcal{D}_{\text{reward}} \;=\; \{(m,\,x,\,\hat{y}_{\text{img}},\,a)\}.$
We stress that GPT-5-mini is used \emph{only} to label the training set $\mathcal{D}_{\text{reward}}$ and plays no role at evaluation time (Sec.~\ref{sec:rewardbench}).
The teacher therefore acts as a scalable labeler for training only, not as both annotator and judge; any residual annotator bias in $\mathcal{D}_{\text{reward}}$ is bounded by \methodname's downstream RL gains (Sec.~\ref{sec:visual_erm_rl}), which are measured on benchmarks independent of the training labels.

\noindent \textbf{Training objective.}
\methodname is a conditional generator $f_{\theta_{\text{ERM}}}(a\mid x,\hat{y}_{\text{img}})$ that produces the structured discrepancy list and per-error severities used in Sec.~\ref{subsec:method_rl}.
We supervise it by negative log-likelihood:
\begin{equation}
\mathcal{L}(\theta_{\text{ERM}})
=
\mathbb{E}_{(m,x,\hat{y}_{\text{img}},a)\sim \mathcal{D}_{\text{reward}}}
\Big[
-\log f_{\theta_{\text{ERM}}}\big(a \mid x,\,\hat{y}_{\text{img}}\big)
\Big].
\end{equation}
For an annotation sequence $a=(a_1,\ldots,a_T)$, this expands to the token-level form
\begin{equation}
\mathcal{L}(\theta_{\text{ERM}})
=
\mathbb{E}_{(m,x,\hat{y}_{\text{img}},a)}
\Big[
-\sum_{t=1}^{T}
\log f_{\theta_{\text{ERM}}}\big(a_t \mid x,\,\hat{y}_{\text{img}},\,a_{<t}\big)
\Big],
\end{equation}
where $a_{<t}=(a_1,\ldots,a_{t-1})$.
A single $f_{\theta_{\text{ERM}}}$ is trained on all $m$, supporting the task-agnostic property in Sec.~\ref{sec:analysis}.

\subsection{\methodname as an RL Reward}
\label{subsec:method_rl}
\noindent \textbf{Pipeline.}
Following Fig.~\ref{fig:framework}(b), each rollout draws $y\sim\pi_{\theta}(\cdot\mid x)$, renders $\hat{y}_{\text{img}}=\mathcal{R}_m(y)$, and queries \methodname on $(x,\hat{y}_{\text{img}})$ to obtain a discrepancy set $\mathcal{E}=\{e_k\}_{k=1}^{K}$ with severities $s_k\!\ge\!0$.
We pair this with a render-success reward (RSR) so that unrenderable code is excluded before fidelity is scored.

\noindent \textbf{Reward design.}\label{sec:reward_design}
We aggregate severities into $\mathrm{S}_{\textsc{verm}} \;=\; \sum_{k=1}^{K} s_k$,
normalize within the task batch $\mathcal{T}$:
\begin{equation}
\widetilde{\mathrm{S}}_{\textsc{verm}}
\;=\;
\frac{\mathrm{S}_{\textsc{verm}}}{\max_{j\in\mathcal{T}} \mathrm{S}_{\textsc{verm}}^{(j)} + \epsilon},
\end{equation}
and convert to a bounded reward
\begin{equation}
r_{\textsc{verm}}
\;=\;
\mathrm{clip}\!\big(1-\widetilde{\mathrm{S}}_{\textsc{verm}},\,0,\,1\big).
\end{equation}
The overall RL reward is
\begin{equation}\label{eq:reward}
    r \;=\; r_{\textsc{rsr}} \;+\; r_{\textsc{verm}},
\end{equation}
where $r_{\textsc{rsr}}=1$ if $\mathcal{R}_m(y)$ succeeds and $0$ otherwise.

\noindent \textbf{Policy optimization.}\label{sec:optimization_rl}
We optimize $\pi_{\theta}$ with a Group Relative Policy Optimization (GRPO)-based~\cite{grpo} objective using the reward in Eq.~\ref{eq:reward} and a Kullback--Leibler (KL) anchor to a reference policy $\pi_{\text{ref}}$:
\begin{equation}
\begin{aligned}
\max_{\theta}\;\;
\mathbb{E}_{x\sim \mathcal{D}}
\Big[
&\mathbb{E}_{y \sim \pi_{\theta}(\cdot \mid x)}
\big[
r_{\textsc{rsr}}(\mathcal{R}_m(y))
\\
& +
\mathrm{clip}\!\big(1-\widetilde{S}_{\textsc{verm}}(x,\,\mathcal{R}_m(y)),\,0,\,1\big)
\big]
\\
&-\beta\,\mathrm{KL}\!\left(\pi_{\theta}(\cdot\mid x)\,\|\,\pi_{\text{ref}}(\cdot\mid x)\right)
\Big],
\end{aligned}
\label{eq:rl_objective}
\end{equation}
where $\beta$ controls the strength of KL regularization and $\mathcal{D}$ is the training distribution over reference images $x$.
The training and downstream behavior of policies optimized under Eq.~\ref{eq:rl_objective} are reported in Fig.~\ref{fig:analysis} and Sec.~\ref{sec:experiments}.

\subsection{\methodname for Test-Time Scaling}
A second use of \methodname, shown in Fig.~\ref{fig:framework}(b), is to provide feedback for iterative self-refinement at inference time.
The policy first produces an initial prediction
\begin{equation}
y^{(0)} \sim \pi_{\theta}(\cdot \mid x),
\end{equation}
which is rendered into $\hat{y}_{\text{img}}^{(0)}=\mathcal{R}_m(y^{(0)})$ and evaluated by \methodname:
\begin{equation}
    \big(r^{(0)},\, f^{(0)}\big) \;=\; f_{\theta_{\text{ERM}}}\!\left(x,\,\hat{y}_{\text{img}}^{(0)}\right),
\end{equation}
where $r^{(0)}$ is the bounded reward from Eq.~\ref{eq:reward} and $f^{(0)}$ is the structured discrepancy description.
If $r^{(0)}$ falls below a threshold, the policy revises its solution conditional on the previous draft and the feedback,
\begin{equation}
y^{(t+1)} \sim \pi_{\theta}\!\left(\cdot \mid x,\, y^{(t)},\, f^{(t)}\right),
\end{equation}
and the loop repeats for up to $T$ steps.
The interpretability of $f^{(t)}$, rather than $r^{(t)}$ alone, is what enables this revision step; a scalar-only reward could not localize \emph{what} to fix.
We evaluate this capability in Sec.~\ref{sec:ablation_test_time_scaling}; gains are bounded by the policy's editing ability, since TTS reuses $\pi_{\theta}$ without weight updates.

\begin{figure*}[t]
    \begin{center}
    \includegraphics[width=1.\linewidth]{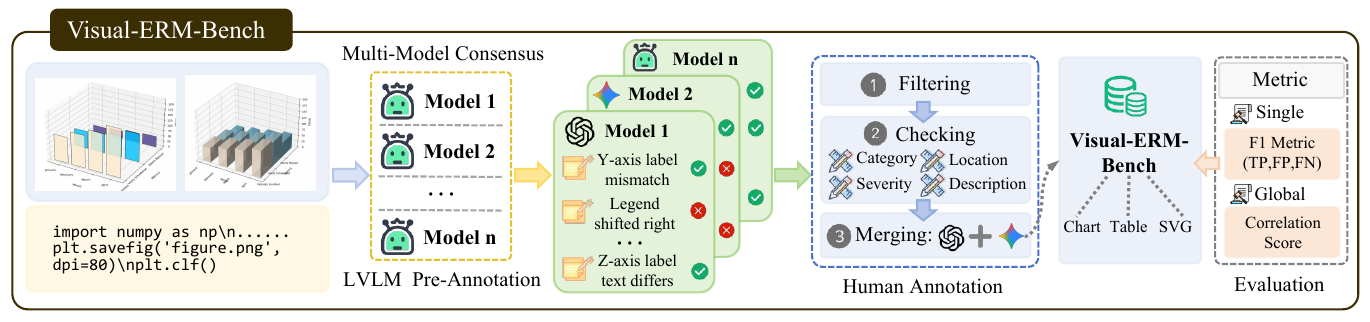}
    \end{center}
    \vspace{-12pt}
    \caption{\small \textbf{\benchname.} We construct \benchname by first leveraging several advanced proprietary models for preliminary annotation, followed by manual consolidation and filtering, resulting in 1,335 high-quality instances.}
    \label{fig:bench}
    \vspace{-14pt}
\end{figure*}

\subsection{VisualCritic-RewardBench}\label{sec:rewardbench}
Existing reward benchmarks predominantly evaluate vision--language alignment~\cite{li2025vl}, whereas the bottleneck for vision-to-code is image-to-image reconstruction fidelity.
We therefore introduce VisualCritic-RewardBench (\benchname), a diagnostic suite that targets fine-grained image-to-image discrepancies in structured visuals.

\benchname contains 1{,}335 high-quality annotated instances, constructed following Fig.~\ref{fig:bench}.
Each instance pairs a reference image $x$ with a corrupted counterpart $\hat{y}_{\text{img}}$ and a fine-grained discrepancy annotation $a$ covering type, location, description, and severity.
We adopt a multi-model consensus protocol: independent annotations from GPT-5-mini~\cite{singh2025openai}, Gemini-2.5-Pro, and Gemini-3-Pro~\cite{comanici2025gemini} reduce architecture-specific bias, and PhD-level reviewers consolidate the labels into a single expert-vetted set.
Example cases are provided in Sec.~\ref{sec:case_study}.

Because \benchname mixes structured fields (e.g., counts) with free-form content (e.g., descriptions), exact-match accuracy is unsuitable.
We adopt an LLM-as-Judge protocol: a judge LLM matches predicted discrepancies $\hat{A}$ to ground-truth annotations $A^{\star}$ to identify TP/FP/FN and compute Precision/Recall/F1.
Since models also output per-discrepancy severities, we sum severities per instance and report the Pearson correlation with the ground-truth totals to measure overall scoring consistency, denoted as $S_c$.
\benchname is a diagnostic for $f_{\theta_{\text{ERM}}}$ itself, separate from the downstream RL and TTS evaluations in Sec.~\ref{sec:experiments}.
\section{Experiments}\label{sec:experiments}
We evaluate \methodname along three axes. (i) \emph{RL utility}: whether using \methodname as the reward in RL yields stronger policies across three vision-to-code tasks (Sec.~\ref{sec:visual_erm_rl}). (ii) \emph{Reward quality}: how well \methodname judges fine-grained image-to-image discrepancies on \benchname (Sec.~\ref{sec:visual_erm_bench_results}). (iii) \emph{Test-time utility}: whether \methodname's critiques improve on test-time scaling (Sec.~\ref{sec:ablation_test_time_scaling}).

\subsection{Reinforcement Learning with Visual-ERM}~\label{sec:visual_erm_rl}

\noindent \textbf{Experimental Details and Benchmarks.}
We train \methodname on top of Qwen3-VL-8B-Instruct~\cite{Bai2025Qwen3VLTR}. To evaluate its utility as a reward model, we run GRPO~\cite{grpo} on three vision-to-code tasks: (1) \textit{Chart-to-Code}, (2) \textit{Table-to-Markdown}, and (3) \textit{SVG-to-Code}, using Qwen3-VL-8B-Instruct as the policy model. 
For \textit{Chart-to-Code}, we use \textsc{ChartMimic}~\cite{yang2024chartmimic} under both the \textit{direct} (reproduce the input chart) and \textit{customized} (generate a new chart under given style/data constraints) settings. For \textit{Table-to-Markdown}, we report table-level metrics on \textsc{OmniDocBench-v1/v1.5}~\cite{ouyang2025omnidocbench} and \textsc{olmOCRBench}~\cite{poznanski2025olmocr}. For \textit{SVG-to-Code}, we evaluate on \textsc{UniSVG}~\cite{li2025unisvg}.
We additionally compare against two stronger vision-to-code SFT policies, VinciCoder~\cite{zhao2025vincicoder} and JanusCoderV-8B~\cite{sun2025januscoder}.
Please refer to Sec.~\ref{sec:appendix_dataset_benchmark_statistic} for more experimental details.

\begin{table*}[t]
\caption{\small \textbf{Evaluation Results on Chart-to-Code Tasks.} We evaluate on ChartMimic(Direct) and ChartMimic(Customized). ChartMimic(Direct) requires generating Python code that reproduces the input chart, while ChartMimic(Customized) requires generating code for a new chart that matches the given chart's style and data constraints. \textbf{Exec\_rate} denotes the execution success rate; \textbf{Low} and \textbf{High} denote the low-level and high-level scores, respectively.}
\vspace{-6pt}
\label{tab:chartmimic_eval}
\centering
\setlength{\tabcolsep}{8pt}         
\renewcommand{\arraystretch}{1.}    
\resizebox{0.95\linewidth}{!}{%
\begin{tabular}{@{} l *4{c} *4{c} c @{}}
\toprule[1.25pt]
\multirow{2}{*}{\textbf{Model}} 
& \multicolumn{4}{c}{\textbf{ChartMimic~\cite{yang2024chartmimic} (Direct)}} & \multicolumn{4}{c}{\textbf{ChartMimic~\cite{yang2024chartmimic} (Customized)}} & \multirow{2}{*}{\textbf{Avg}$\uparrow$} \\
\cmidrule(lr){2-5}\cmidrule(lr){6-9}
& Exec\_rate$\uparrow$ & Low$\uparrow$ & High$\uparrow$ & Overall$\uparrow$ & Exec\_rate$\uparrow$ & Low$\uparrow$ & High$\uparrow$ & Overall$\uparrow$ & \\
\midrule
\addlinespace[2pt]
\textit{Baseline} \\[-2pt]
\addlinespace[2pt]
InternVL3.5-8B~\cite{wang2025internvl3} & 62.2 & 41.9 & 48.9 & 45.4 & 79.7 & 53.0 & 63.3 & 58.2 & 51.8 \\
Qwen3-VL-8B-Instruct~\cite{Qwen3-VL} & 82.2 & 62.7 & 72.7 & 67.7 & 86.8 & 66.3 & 76.8 & 71.6 & 69.6 \\
VinciCoder-8B-SFT~\cite{zhao2025vincicoder} & 86.2&	68.2&	77.5&	72.9&	86.2&	52.3&	71.5&	61.9&	67.4  \\
JanusCoderV-8B~\cite{sun2025januscoder} & 80.6&	65.8&	73.2&	69.5&	80.7&	66.7&	74.2&	70.4&	70.0  \\
ChartMaster-7B~\cite{tan2025chartmaster} & 93.3&	74.5&	82.1&	78.3&	88.8&	59.5&	74.2&	66.8&	72.6  \\
\midrule
\textit{Qwen3-VL-8B-Instruct~\cite{Qwen3-VL}} \\[-2pt]
\addlinespace[2pt]
$+$ RL (DINO-based Reward) &  90.8&	71.1&	81.9&	76.5&	91.7&	71.2&	80.3&	75.8&	76.1  \\
\rowcolor[HTML]{DAEFF9}
$+$ RL (Visual-ERM) & 92.5 & 74.4 & 84.6 & 79.5 & 91.5 & 71.8 & 81.1 & 76.5 & 78.0 \\
$\Delta_{\text{vs.\ Qwen3-VL-8B-Instruct}}$  & \hgreen{+10.3} & \hgreen{+11.9} & \hgreen{+11.9} & \hgreen{+11.8} & \hgreen{+4.7} & \hgreen{+5.5} & \hgreen{+4.3} & \hgreen{+4.9} & \hgreen{+8.4} \\
\midrule
\textit{VinciCoder-8B-SFT~\cite{zhao2025vincicoder}} \\[-2pt]
\addlinespace[2pt]
$+$ RL (DINO-based Reward) & 90.8&	73.1&	82.3&	77.7&	91.5&	57.0&	77.5&	67.3&	72.5  \\
\rowcolor[HTML]{DAEFF9} 
$+$ RL (Visual-ERM) & 94.3&	78.0&	88.5&	83.2&	95.0&	60.9&	82.6&	71.7&	77.5  \\
$\Delta_{\text{vs.\ VinciCoder-8B-SFT}}$  & \hgreen{+8.1} & \hgreen{+9.8} & \hgreen{+11.0} & \hgreen{+10.3} & \hgreen{+8.8} & \hgreen{+8.6} & \hgreen{+11.1} & \hgreen{+9.8} & \hgreen{+10.1} \\
\bottomrule[1.25pt]
\end{tabular}%
}
\vspace{-12pt}
\end{table*}

\begin{table*}[t]
\caption{\small \textbf{Evaluation Results on Table-to-Markdown.} TEDS-S denotes TEDS-Structure-Only, and TA represents the table subtask of olmOCRBench. For Edit-Dist, where lower values indicate better performance, we use ($100 - \text{Edit-Dist}$) when computing the Average.}
\vspace{-6pt}
\label{tab:table_to_markdown_rl}
\centering
\setlength{\tabcolsep}{8pt}         
\renewcommand{\arraystretch}{1.}    
\resizebox{0.95\linewidth}{!}{%
\begin{tabular}{@{} l *3{c} *3{c} c c @{}}
\toprule[1.25pt]
\multirow{2}{*}{\textbf{Model}} 
& \multicolumn{3}{c}{\textbf{OmniDocBench~\cite{ouyang2025omnidocbench}}} & \multicolumn{3}{c}{\textbf{OmniDocBench-v1.5~\cite{ouyang2025omnidocbench}}} & \multicolumn{1}{c}{\textbf{olmOCRBench~\cite{poznanski2025olmocr}}} & \multirow{2}{*}{\textbf{Avg}$\uparrow$} \\
\cmidrule(lr){2-4}\cmidrule(lr){5-7}\cmidrule(lr){8-8}
& TEDS$\uparrow$ & TEDS-S$\uparrow$ & Edit-Dist$\downarrow$ & TEDS$\uparrow$ & TEDS-S$\uparrow$ & Edit-Dist$\downarrow$ & TA$\uparrow$ & \\
\midrule
\addlinespace[2pt]
\textit{Baseline} \\[-2pt]
\addlinespace[2pt]
Qwen2.5-VL-7B~\cite{Qwen2.5-VL}  & 74.4 & 80.1 & 58.9 & 69.3 & 74.1 & 58.7 & 71.1 & 64.5 \\
InternVL3.5-8B~\cite{wang2025internvl3}  & 71.0 & 77.6 & 44.9 & 62.9 & 70.4 & 44.1 & 74.1 & 66.7 \\ 
Qwen3-VL-8B-Instruct~\cite{Qwen3-VL}  & 78.9 & \underline{83.9} & \underline{23.2} & 72.7 & \underline{77.2} & \underline{26.9} & 75.3 & \underline{76.8}\\
\midrule
\textit{Qwen3-VL-8B-Instruct~\cite{Qwen3-VL}} \\[-2pt]
\addlinespace[2pt]
$+$ RL (DINO-based Reward) & 62.2 & 69.5 & 37.0 & 61.1 & 67.4 & 37.9 & 71.7 & 65.3 \\
$+$ RL (TEDS-based Reward) & \underline{79.2} & 82.9 & 31.6 & \underline{73.0} & 76.6 & 35.3 & \textbf{78.6} & 74.8 \\
\midrule
\rowcolor[HTML]{DAEFF9} 
$+$ RL (Visual-ERM) & \textbf{81.4} & \textbf{86.3} & \textbf{20.7} & \textbf{75.4} & \textbf{80.4} & \textbf{24.2} & \underline{78.1} & \textbf{79.5} \\
$\Delta_{\text{vs.\ Qwen3-VL-8B-Instruct}}$ & \hgreen{+2.5} & \hgreen{+2.4} & \hgreen{+2.5} & \hgreen{+2.7} & \hgreen{+3.2}  & \hgreen{+2.7} & \hgreen{+2.8} & \hgreen{+2.7} \\
\bottomrule[1.25pt]
\end{tabular}%
}
\vspace{-12pt}
\end{table*}

\noindent \textbf{Results on Chart-to-Code.}
We start with the \textit{Chart-to-Code} task. Using Qwen3-VL-8B-Instruct~\cite{Qwen3-VL} and VinciCoder-8B-SFT~\cite{zhao2025vincicoder} as policies, we run GRPO with \methodname as the reward. Tab.~\ref{tab:chartmimic_eval} shows that \methodname-guided RL improves Qwen3-VL-8B-Instruct by \hgreen{+11.8} and \hgreen{+4.9} average points on ChartMimic-v2 \textit{direct} and \textit{customized}~\cite{yang2024chartmimic}. Starting from the already strong VinciCoder-8B-SFT, the same recipe still delivers \hgreen{+10.3} and \hgreen{+9.8} under the two settings.

We also compare \methodname against the DINO-based RL reward in Tab.~\ref{tab:chartmimic_eval}, and observe three advantages. \emph{(1) Stronger downstream policies}: \methodname-guided RL yields substantially larger gains than DINO-based RL on both Qwen3-VL-8B-Instruct and VinciCoder-8B-SFT backbones. DINO rewards reduce supervision to a fixed visual embedding space, where matching patch-level feature similarity prioritizes semantic alignment and global appearance while under-penalizing small but functionally critical deviations, so the proxy only loosely tracks human-perceived fidelity. \emph{(2) Cross-modal coverage}: DINO rewards are unimodal and weak at penalizing errors carried by rendered text, allowing policies to improve the proxy while degrading textual faithfulness; \methodname combines visual perception with cross-modal grounding, scoring reconstructions on both visual structure and rendered text. \emph{(3) Fine-grained interpretable feedback}: trained as a generative judge, \methodname emits decomposed discrepancy descriptions rather than a single scalar, which directly supports test-time scaling via reward-guided refinement (Sec.~\ref{sec:ablation_test_time_scaling}).

\noindent \textbf{Results on Table-to-Markdown Parsing.}
We next turn to the \textit{Table-to-Markdown} task. Starting from Qwen3-VL-8B-Instruct, we run GRPO with \methodname as the reward and compare against two alternative signals: a rule-based Tree-Edit-Distance Similarity (TEDS) score, and a DINO-based feature similarity. Tab.~\ref{tab:table_to_markdown_rl} shows that \methodname-based RL delivers consistent improvements, with an overall gain of \hgreen{+2.7}; the gains span both textual-recognition and structural-reconstruction metrics, consistent with the cross-modal coverage of \methodname's feedback.

Both alternatives fall short on this task. With TEDS, the training reward rises steadily, yet the policy improves only marginally on the target TEDS metric and slightly degrades on others, a pattern consistent with reward shortcutting in a purely textual/structural space that ignores visual cues. DINO-based RL fares worse: unlike charts, where DINO features still carry useful visual cues, tables are dominated by precise text and layout, and DINO-based RL not only fails to yield gains on \textit{Table-to-Markdown} but also degrades performance across benchmarks.

\begin{table*}[t]
\vspace{-4pt}
\centering

\begin{minipage}[t]{0.49\linewidth}
\vspace{0pt}
\centering
\captionof{table}{\small \textbf{Results on UniSVG.} SSIM/LPIPS measure pixel-level and perceptual similarity, CLIP measures semantic alignment, and \textbf{Score} aggregates them.}
\vspace{-6pt}
\setlength{\tabcolsep}{2pt}
\renewcommand{\arraystretch}{1.}
\resizebox{.9\linewidth}{!}{%
\begin{tabular}{@{} l *4{c} @{}}
\toprule[1.25pt]
\multirow{2}{*}{Model}
& \multicolumn{4}{c}{\textbf{UniSVG~\cite{li2025unisvg} (ISVGEN)}} \\
\cmidrule(lr){2-5}
& SSIM$\uparrow$ & LPIPS$\downarrow$ & CLIP$\uparrow$ & \textbf{Score}$\uparrow$ \\
\midrule
\textit{Baseline} \\[-2pt]
Qwen3-VL-8B        & 60.9 & 60.0 & 73.3 & 64.2 \\
JanusCoderV-8B     & 58.1 & 61.7 & 72.6 & 62.8 \\
\midrule
VinciCoder-8B-SFT  & 81.1 & 19.2 & 92.5 & 87.9 \\
$+$ RL (DINO)      & 76.9 & 23.3 & 92.7 & 86.3 \\
\rowcolor[HTML]{DAEFF9}
$+$ RL (\methodname) & 85.2 & 12.6 & 95.2 & 91.6 \\
\bottomrule[1.25pt]
\end{tabular}%
}
\label{tab:svg_to_code}
\end{minipage}\hfill
\hspace{+2pt}
\begin{minipage}[t]{0.49\linewidth}
\centering
\captionof{table}{\small \textbf{Test-Time Scaling on \textit{Chart-to-Code}.} \methodname enables iterative self-reflection and revision; for the reflection turns study, see Tab.~\ref{tab:ablation_reflection_turns}.}
\vspace{-6pt}
\setlength{\tabcolsep}{3pt}
\renewcommand{\arraystretch}{1.}
\resizebox{.9\linewidth}{!}{%
\begin{tabular}{@{} l c c c @{}}
\toprule[1.25pt]
\multirow{2}{*}{Model}
& \multicolumn{2}{c}{\textbf{ChartMimic~\cite{yang2024chartmimic}}} & \multirow{2}{*}{\textbf{Avg}$\uparrow$} \\
\cmidrule(lr){2-3}
& Direct$\uparrow$ & Cust.$\uparrow$ & \\
\midrule
Qwen3-VL-8B-Instruct~\cite{Qwen3-VL}          & 67.7 & 71.6 & 69.6 \\
$+$ Reflection (self)         & 61.9 & 69.5 & 65.7 \\
\rowcolor[HTML]{DAEFF9}
$+$ Reflection (\methodname)  & 75.6 & 79.5 & 77.6 \\
\midrule
$+$ RL (\methodname)          & 79.5 & 76.5 & 78.0 \\
$+$ Reflection (self)         & 75.2 & 76.9 & 76.1 \\
\rowcolor[HTML]{DAEFF9}
$+$ Reflection (\methodname)  & 80.3 & 82.0 & 81.1 \\
\bottomrule[1.25pt]
\end{tabular}%
}
\label{tab:reflection_on_chart}
\end{minipage}
\vspace{-8pt}
\end{table*}

\begin{table*}[t]
\caption{\small \textbf{Evaluation Results on \benchname.} We evaluate a range of proprietary and open-source models on Visual-ERM-Bench. $F1_h$ denotes $F1_{hard}$, the strict-match F1 score; $F1_s$ denotes $F1_{soft}$, the soft-match F1 score; and $S_c$ denotes the correlation score, measuring the overall agreement between the predicted scores and the ground-truth labels.}
\vspace{-6pt}
\label{tab:visual_erm_bench}
\centering
\setlength{\tabcolsep}{8pt}         
\renewcommand{\arraystretch}{.95}    
\resizebox{0.97\linewidth}{!}{%
\begin{tabular}{@{} l | *3{c} | *3{c} | *3{c} | *3{c} @{}}
\toprule[1.25pt]
\multirow{2}{*}{\textbf{Model}} 
& \multicolumn{3}{c}{\textbf{Chart}} & \multicolumn{3}{|c}{\textbf{Table}} & \multicolumn{3}{|c|}{\textbf{SVG}} & \multicolumn{2}{|c}{\textbf{AVG}} \\
\cmidrule(lr){2-4}\cmidrule(lr){5-7}\cmidrule(lr){8-10}\cmidrule(lr){11-13}
& $F1_h$ & $F1_s$ & $S_c$ & $F1_h$ & $F1_s$ & $S_c$ & $F1_h$ & $F1_s$ & $S_c$ & $F1_h$ & $F1_s$ & $S_c$ \\
\midrule
\rowcolor{gray!20}
\multicolumn{13}{c}{\textit{Proprietary}} \\
\midrule
GPT-4o~\cite{achiam2023gpt4} & 22.8 & 28.3 & 48.5 & 32.9 & 35.7  & 49.5 & 13.0 & 19.3 & 50.3 & 25.0 & 29.5 &  56.5\\
GPT-5.2~\cite{singh2025openai} & 30.1 & 32.6 & 64.8 & 39.3 & 40.6 & 54.6 & 28.5 & 32.2 & 61.1 & 32.7 & 35.0 & 58.9 \\
Gemini-2.5-Pro~\cite{comanici2025gemini}  & 33.7 & 37.5 & 61.8 & 46.4 & 48.0 & 49.9 & 29.3 & 34.3 & 63.3 & 37.8 & 40.9 & 59.1 \\
Gemini-3-Flash~\cite{gemini3.1pro}  & 38.5 & 41.3 & 62.8 & 48.1 & 50.1 & 45.6 & 33.3 & 67.5 & 64.3 & 40.6 & 43.4 & 53.4 \\
\midrule
\rowcolor{gray!20}
\multicolumn{13}{c}{\textit{Open-source}} \\
\midrule
Qwen2.5-VL-7B~\cite{Qwen2.5-VL}  & 3.9 & 5.4 & 11.2 & 2.3 & 3.1 & 12.6 & 1.9 & 7.5 & 37.9 & 2.8 & 5.1 & 15.2 \\
InternVL3.5-8B~\cite{wang2025internvl3}  & 2.5 & 5.7 & 11.0 & 9.9 & 10.9 & 31.7 & 6.1 & 13.1 & 48.9 & 6.7 & 9.6 & 32.5 \\
Qwen3-VL-8B-Instruct~\cite{Qwen3-VL}  & 3.3 & 3.5 & 3.8 & 7.0 & 7.8 & 21.4 & 6.1 & 9.4 & 27.1 & 5.3 & 6.5 & 17.5 \\
Qwen3-VL-235B-Instruct~\cite{Qwen3-VL}  & 28.0 & 31.8 & 47.2 & 35.7 & 37.4 & 56.2 & 19.4 & 22.8 & 51.5 & 29.5 & 32.4 & 56.2 \\
\midrule
\rowcolor[HTML]{DAEFF9} Visual-ERM  & 39.9 & 42.8 & 61.2 & 56.4 & 57.6 & 74.8 & 28.3 & 32.6 & 59.6 & 42.1 & 44.7 & 58.4 \\
$\Delta_{\text{vs.\ Qwen3-VL-8B-Instruct}}$ & \hgreen{+36.6} & \hgreen{+39.3} & \hgreen{+57.4} & \hgreen{+49.4} & \hgreen{+49.8} & \hgreen{+53.4} & \hgreen{+22.2} & \hgreen{+23.2} & \hgreen{+32.5} & \hgreen{+36.8} & \hgreen{+38.2} & \hgreen{+40.9} \\
\bottomrule[1.25pt]
\end{tabular}%
}
\vspace{-12pt}
\end{table*}

\noindent \textbf{Results on SVG-to-Code Parsing.}
We use VinciCoder-8B-SFT as policy and DINO-based RL as the feature-similarity baseline. Compared with charts and tables, SVG reconstruction relies more on visual geometry and styling cues and less on textual content.
Tab.~\ref{tab:svg_to_code} shows that \methodname-based RL delivers consistent gains on the backbone. Notably, while DINO-based RL leads to performance degradation on the strong VinciCoder-8B-SFT baseline, \methodname maintains its effectiveness. This suggests that \methodname provides more robust and precise guidance than standard feature-similarity rewards, which may fail to provide meaningful gradients for highly optimized policies.

\subsection{Reward-Model Evaluation on \benchname}\label{sec:visual_erm_bench_results}

In Sec.~\ref{sec:visual_erm_rl}, the selected benchmarks measure vision-to-code parsing quality but do not directly probe whether a model can judge reconstruction fidelity or surface fine-grained discrepancies. We therefore evaluate on \benchname, which targets fine-grained discrepancy detection and interpretable feedback across vision-to-code tasks.
Tab.~\ref{tab:visual_erm_bench} reports the results on \benchname. Built on Qwen3-VL-8B-Instruct, \methodname improves $F1_h$/$F1_s$/$S_c$ over its base by \hgreen{+36.8}/\hgreen{+38.2}/\hgreen{+40.9}. Even Qwen3-VL-235B-Instruct struggles with fine-grained visual and textual discrepancies, whereas \methodname at 8B matches or surpasses the strongest proprietary baselines on $F1_h$/$F1_s$ and remains competitive on $S_c$. We read this as evidence that fine-grained discrepancy detection and fidelity judgment come from reward-model specialization of a general-purpose LVLM, not from scale alone.

\subsection{Ablation Studies}\label{sec:ablation_study}

\begin{figure}[t]
    \centering
    \includegraphics[width=\linewidth]{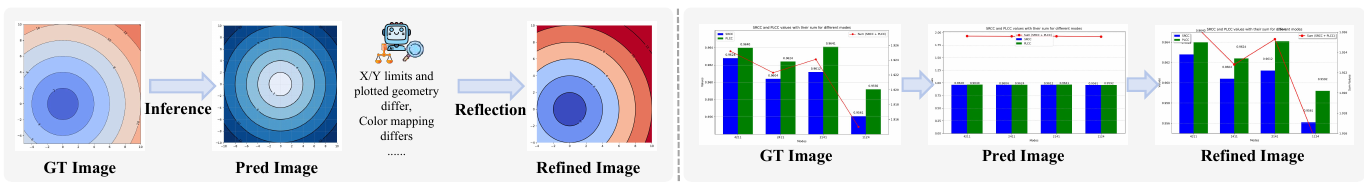}
    \vspace{-12pt}
    \caption{\small \textbf{Test-Time Scaling Cases.} Using \methodname's fine-grained feedback, we enable an inference, reflection and refinement loop. Predictions are generated as text and rendered as images for visualization.}
    \vspace{-12pt}
    \label{fig:chart_reflection_case}
\end{figure}

\noindent \textbf{Test-Time Scaling with Visual-ERM.}\label{sec:ablation_test_time_scaling}
\methodname's feedback is also useful at inference time. Rather than emitting only a scalar score, it returns localized discrepancy descriptions that a policy can act on. We insert \methodname into the decoding loop: it critiques each candidate, and the policy revises its output before finalizing.
With three reflection rounds on \textit{Chart-to-Code}, this loop adds \hgreen{+8.0} Avg over the base Qwen3-VL-8B-Instruct and a further \hgreen{+3.1} on top of the \methodname-RL-tuned policy (Tab.~\ref{tab:reflection_on_chart}). Qualitative cases are provided in Fig.~\ref{fig:chart_reflection_case}.

\noindent \textbf{Ablation of the Reflection Rounds in Test-Time Scaling.}\label{sec:ablation_reflection_turns}
In Tab.~\ref{tab:reflection_on_chart}, we use a three-round reflection and revision pipeline by default. To study how the number of reflection rounds affects performance, we conduct an ablation where we vary the number of rounds while keeping the rest of the inference setup identical. Results are summarized in Tab.~\ref{tab:ablation_reflection_turns}. Overall, increasing the number of reflection rounds yields consistent improvements, with diminishing returns beyond three rounds.

\begin{table*}[t]
\caption{\small \textbf{Ablation on the number of reflection rounds in test-time scaling.} We evaluate test-time scaling with different numbers of reflection and revision rounds d.}
\vspace{-6pt}
\label{tab:ablation_reflection_turns}
\centering
\setlength{\tabcolsep}{4pt}         
\renewcommand{\arraystretch}{1.}    
\resizebox{1.\linewidth}{!}{%
\begin{tabular}{@{} l *4{c} *4{c} c @{}}
\toprule[1.25pt]
\multirow{2}{*}{\textbf{Model}} 
& \multicolumn{4}{c}{\textbf{ChartMimic~\cite{yang2024chartmimic} (Direct)}} & \multicolumn{4}{c}{\textbf{ChartMimic~\cite{yang2024chartmimic} (Customized)}} & \multirow{2}{*}{\textbf{Avg}$\uparrow$} \\
\cmidrule(lr){2-5}\cmidrule(lr){6-9}
& exec\_rate$\uparrow$ & average$\uparrow$ & gpt\_score$\uparrow$ & overall$\uparrow$ & exec\_rate$\uparrow$ & average$\uparrow$ & gpt\_score$\uparrow$ & overall$\uparrow$ & \\
\midrule
Qwen3-VL-8B-Instruct~\cite{Qwen3-VL} (+RL on Visual-ERM) & 92.5 & 74.4 & 84.6 & 79.5 & 91.5 & 71.8 & 81.1 & 76.5 & 78.0 \\
\midrule
$+$ Reflection ($w$ Visual-ERM) 2 rounds & 92.8 & 75.2 & 84.5 & 79.8 & 92.0 & 74.4 & 82.8 & 78.6 & 79.2 \\
\rowcolor[HTML]{DAEFF9}
$+$ Reflection ($w$ Visual-ERM) 3 rounds & 92.3 & 74.8 & 85.8 & 80.3 & 94.7 & 77.1 & 86.8 & 82.0 & 81.1 \\
$+$ Reflection ($w$ Visual-ERM) 4 rounds & 92.0 & 74.5 & 85.4 & 80.0 & 94.2 & 76.9 & 86.1 & 81.5 & 80.7 \\
\addlinespace[2pt]
\bottomrule[1.25pt]
\end{tabular}%
}
\end{table*}

\noindent \textbf{Additional analysis.} We provide extended results in the appendix, such as an evaluation on general VQA benchmarks to verify that \methodname-guided RL preserves broad multimodal competence (Sec.~\ref{sec:appendix_general_vqa_benchmark}), an ablation on multi-task data mixing for both reward modeling (Sec.~\ref{sec:data_mixing}) and downstream RL (Sec.~\ref{sec:data_mixing_rl}), a robustness study across LLM judges on \benchname (Sec.~\ref{sec:appendix_judge_model_ablation}), and an ablation on reward design with and without the render-success term (Sec.~\ref{sec:appendix_reward_design}).
\section{Related Work}

\noindent \textbf{Reward Models.}
To enable effective RL~\cite{grpo,liu2025visual}, reward models (RMs) provide feedback that guide policy optimization. RMs can take several forms: (1) \textbf{Bradley--Terry (BT)} models that learn a scalar reward from pairwise comparisons and are often instantiated as discriminative rankers~\cite{Cai2024InternLM2TR, starling2023, xcomposer2.5-reward}; (2) \textbf{generative} RMs that produce natural language critiques or judgments which can be mapped to rewards~\cite{Kim2023SOLAR1S, Yuan2024SelfRewardingLM,wang2025unified,liu2025spark}; and (3) \textbf{thinking/agentic} RMs that perform multi-step evaluation, e.g., decomposing criteria, self-reflection, or invoking tools before returning a final score~\cite{ding2025arm,li2025one,peng2025agentic}.
Most prior RMs are developed for text-centric generation (e.g., writing and dialogue) and do not support \emph{visual-to-code} tasks, where quality is mainly determined by visual fidelity rather than text. Therefore, we propose \methodname, a visual equivalence reward model for visual-to-code tasks.

\noindent \textbf{Visual-to-Code Tasks.}
Visual-to-code spans a family of practical structured perception tasks that convert images into executable or structured representations. \emph{Chart-to-Code} aims to parse charts into Python programs that can faithfully reproduce the original plots~\cite{zhao2025chartcoder,tan2025chartmaster,zhang2025enhancing}. \emph{Table-to-Markdown} converts tabular images into structured formats such as Markdown or HTML~\cite{ling2025table2latex_doc_rl,zhang2025monkeyocr_doc_rl,niu2025mineru2}. \emph{SVG-to-Code} translates vector graphics into code representations~\cite{li2025unisvg,yang2025omnisvg}. Such structured outputs facilitate downstream use and improve usability in real-world applications.

\noindent \textbf{RL for Visual-to-Code Tasks.}
Despite its practical importance, visual-to-code remains challenging. Supervised fine-tuning (SFT) typically relies on large-scale, high-quality datasets~\cite{zhao2025chartcoder, zhong2019image, gui2025webcode2m}, which are costly to curate. RL has been explored as an alternative, yet existing reward designs often fall into two extremes: (i) textual rule-based rewards~\cite{ling2025table2latex_doc_rl}, which score string-level or structural proxies in the text space without directly leveraging the visual evidence, and thus may introduce modality bias; and (ii) visual-encoder similarity-based rewards, such as DINO-based~\cite{simeoni2025dinov3} similarity~\cite{zhao2025vincicoder,tan2025chartmaster,rodriguez2025rendering}, which compare representations extracted by vision encoders but are often coarse-grained and offer limited interpretability. Motivated by these limitations, we propose \methodname, a cross-modal reward model that provides fine-grained, interpretable, and task-agnostic feedback for Visual-to-Code.


\vspace{-4pt}
\section{Conclusion}
\vspace{-4pt}

We propose the Visual Equivalence Reward Model (\methodname), a generative reward model that evaluates vision-to-code outputs in visual space, providing fine-grained, interpretable, and task-agnostic supervision. We further introduce VisualCritic-RewardBench to directly assess image-to-image discrepancy judgment across vision-to-code tasks. Experiments show that \methodname is an effective supervisor for both reinforcement learning and test-time scaling, consistently improving vision-to-code performance across multiple tasks.

\clearpage
\bibliographystyle{plain}
\bibliography{refs}


\clearpage
\appendix
\newpage
\appendix
\onecolumn

\section*{Appendix}
In the appendix, we provide additional materials to support the main paper and facilitate a deeper understanding of \methodname.  

First, in Sec.~\ref{sec:appendix_dataset_benchmark_statistic}, we provide detailed descriptions of the models, datasets, benchmark construction, experimental setups and computational efficiency throughout the paper, aiming to ensure reproducibility.

Second, Sec.~\ref{sec:appendix_more_experiments} presents additional experimental analyses and ablation studies, including evaluations on general VQA benchmarks, investigations of multi-task reward modeling, robustness analyses of different judge models, further studies on reward design, test-time scaling and so on.  

Third, in Sec.~\ref{sec:appendix_prompts}, we list the full prompt templates used in our experiments, covering reward model data generation, \methodname inference across different vision-to-code tasks, and the evaluation protocol of VC-RewardBench.  

Finally, Sec.~\ref{sec:case_study} provides representative qualitative case studies from chart-to-code, table-to-markdown, and SVG-to-code tasks, illustrating typical visual discrepancies and the fine-grained feedback produced by \methodname.

\section{More Details}
\label{sec:appendix_dataset_benchmark_statistic}

\subsection{Models}
\label{sec:appendix_models}

This work involves multiple models of different roles and scales. For clarity and reproducibility, we summarize the models used throughout the paper and their corresponding usage in each stage of the pipeline.

\paragraph{\methodname (Reward Model).}
We build \methodname on top of Qwen3-VL-8B-Instruct~\cite{Qwen3-VL}. While Qwen3-VL-8B-Instruct already exhibits strong multimodal perception and reasoning abilities, we find that it is not a reliable \emph{image-to-image discrepancy} judge out of the box—especially for structured visuals where the key differences are often text- and layout-centric. This limitation is empirically reflected in Sec.~\ref{sec:visual_erm_bench_results}, where the base model shows poor discrimination on fine-grained visual mismatches.

Starting from this 8B backbone, we train \methodname on a large-scale reward-modeling corpus spanning three vision-to-code domains: Chart-to-Code (104K), Table-to-Markdown (125K), and SVG-to-Code (111K). This training equips \methodname with task-agnostic yet fine-grained judgment capabilities. Notably, despite its relatively small size, \methodname substantially outperforms Qwen3-VL-235B-Instruct as a judge and achieves performance competitive with strong proprietary models, demonstrating that targeted reward model training can be more effective than simply scaling up a general-purpose LVLM.

\paragraph{Policy models for RL.}
For reinforcement learning, we consider two types of policy backbones. The first is Qwen3-VL-8B-Instruct, which serves as our default policy model across tasks. In this setting, we use \methodname as the reward model and optimize Qwen3-VL-8B-Instruct with RL to improve its vision-to-code generation quality.

In addition, to test whether \methodname can also improve stronger specialized parsers, we use VinciCoder-8B-SFT~\cite{zhao2025vincicoder} as an alternative policy backbone on Chart-to-Code and SVG-to-Code. VinciCoder-8B-SFT provides a stronger supervised baseline for these tasks, and applying \methodname-guided RL on top of it allows us to evaluate the reward model’s transferability and its ability to refine already-competitive policies.

For Table-to-Markdown, we do not include VinciCoder-8B-SFT because it does not have a comparable task-specific SFT baseline for tables; therefore, we only use Qwen3-VL-8B-Instruct as the policy model for RL in this setting.

\subsection{Datasets}
\label{sec:appendix_datasets}

We use multiple datasets throughout the paper for reward-model training, RL policy optimization, and benchmark construction. For completeness, we summarize them here. The annotations and training data will be released publicly in the future, and qualitative examples are provided in Sec.~\ref{sec:case_study}.

\paragraph{Reward-model training data.}\label{appd:reward_data_gen}

\begin{figure}[t]
    \centering
    \includegraphics[width=\linewidth]{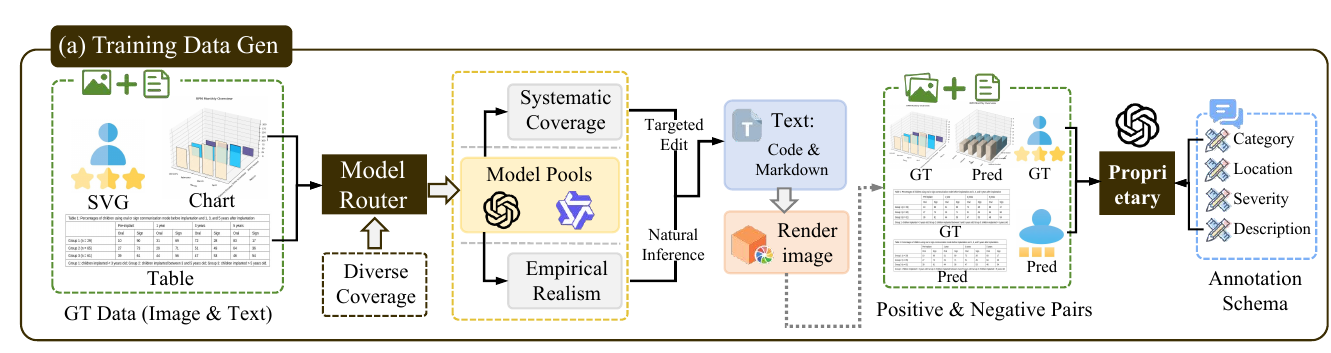}
    \vspace{-12pt}
    \caption{\small \textbf{Training Data Generation.} We construct our \methodname's training data as shown in Figure.}
    \vspace{-12pt}
    \label{fig:train_data_gen}
\end{figure}

We first collect a pool of images and their corresponding ground-truth annotations from public vision-to-code datasets spanning charts, tables, and SVG-style graphics~\cite{zhao2025chartcoder,chen2025breaking,jiang2025viscodex,li2025unisvg,rodriguez2023starvector}. However, these raw pairs are not directly suitable for training a reward model, because \methodname requires supervision on \emph{image-to-image discrepancies} rather than only paired input-output examples. We therefore construct training instances by synthesizing realistic error patterns and converting them into \emph{rendered} visual differences.

Concretely, we create imperfect predictions via two complementary routes: (i) \emph{error injection} on ground-truth annotations using a strong yet cost-effective model (GPT-5-mini) as well as Qwen3-VL-235B-Instruct, and (ii) \emph{model-generated predictions} by running Qwen3-VL-8B-Instruct on a subset of ground-truth images to obtain naturally occurring parsing errors. We then render each prediction back into an image, forming an (Original Image, Generated Image) pair that reflects the types of distortions encountered in practice. Given each image pair, we use GPT-5-mini to produce structured annotations, including fine-grained error descriptions (category, location, severity), as illustrated in Sec.~\ref{sec:case_study}. This procedure yields three reward-modeling datasets: 104K instances for Chart-to-Code, 125K for Table-to-Markdown, and 111K for SVG-to-Code. The trainning data generation pipeline is shown in Fig.~\ref{fig:train_data_gen}.

\paragraph{RL training data.}
The RL data are drawn from the same underlying sources, but require substantially less processing. Unlike reward-model training, RL does not require free-form error annotations: during training, \methodname directly compares the original image with the re-rendered prediction and outputs an image-to-image reward. As a result, we only need the ground-truth images, and in many cases the ground-truth parsing text is not necessary. In our experiments, we use 11K images for Chart-to-Code RL, 40K for Table-to-Markdown RL, and 10K for SVG-to-Code RL.

\paragraph{\benchname construction data.}
We construct \benchname using a similar discrepancy-centric philosophy, but with substantially stricter quality control to ensure reliability as an evaluation benchmark. We begin with a candidate pool of 4.5K image pairs and obtain annotations from three state-of-the-art proprietary models: GPT-5-mini~\cite{singh2025openai}, Gemini2.5-Pro, and Gemini3-Pro~\cite{comanici2025gemini}. We then manually consolidate, filter, and curate the collected annotations to resolve inconsistencies and remove ambiguous or low-quality cases. The final benchmark contains 1,335 carefully vetted examples, which we use to evaluate fine-grained image-to-image judging ability in structured visual domains.

\subsection{\benchname Statistics}
\label{sec:appendix_vc_rewardbench_stats}

This section reports detailed statistics of \benchname. The benchmark contains 1,335 curated examples spanning three structured-visual domains: 595 chart instances, 442 SVG instances, and 298 table instances.

Beyond instance counts, \benchname provides fine-grained error annotations that characterize the discrepancy types between the original image and the re-rendered prediction. For the chart subset, the annotations include 231 \texttt{text\_error}, 619 \texttt{style\_error}, 462 \texttt{data\_error}, and 62 \texttt{structure\_error}. For the table subset, we annotate 884 \texttt{text\_error}, 284 \texttt{numeric\_error}, and 185 \texttt{layout\_error}. For the SVG subset, the error annotations contain 151 \texttt{style\_error}, 218 \texttt{structure\_error}, 360 \texttt{shape\_error}, and 12 \texttt{text\_symbol\_error}.

Overall, these statistics indicate that \benchname covers a diverse range of realistic failure modes. In particular, chart examples are dominated by style- and data-related discrepancies, tables primarily exhibit text-centric recognition errors with a non-trivial portion of numeric and layout issues, and SVG examples emphasize geometric \texttt{shape\_error} and structural composition mismatches. This diversity makes \benchname a challenging and diagnostic benchmark for evaluating fine-grained image-to-image judging capability.

\subsection{Experimental Setups}
\label{sec:appendix_experiment_setups}

We implement supervised training and RL fine-tuning using LLaMA-Factory and the VERL framework. For training \methodname, we use a learning rate of $1\times 10^{-5}$ with a batch size of 32. We train for 3 epochs over the reward-modeling dataset. For reinforcement learning, we adopt GRPO with a learning rate of $1\times 10^{-6}$ and a batch size of 256. For each prompt, we sample 8 rollouts per GRPO update step to estimate the policy gradients.

\subsection{Computational Efficiency and Training Latency}
A primary concern in scaling reinforcement learning (RL) is the computational overhead introduced by reward models, particularly as their parameter count increases. We systematically benchmark the training latency of three reward supervision paradigms under identical experimental configurations: (1) \textbf{Text-based rules} (e.g., TEDS); (2) \textbf{Vision-encoder similarity} (DINOv2-large, 0.3B), following the framework in \cite{zhao2025vincicoder}; and (3) our proposed \methodname (8B).

While text-based rewards are predictably the most efficient at 0.15 hours per step, they yield the least effective policy performance. Notably, \methodname demonstrates a distinct computational advantage: despite its 8B parameter scale, it achieves a per-step latency of 0.17 hours, outperforming the 0.24 hours required by the 0.3B DINOv2-large baseline. This efficiency gap arises because DINO-based rewards~\cite{zhao2025vincicoder}, necessitate segmenting high-resolution images into multiple sub-patches to ensure scoring precision, which introduces significant computational overhead. In contrast, \methodname's unified multimodal architecture processes structured visuals natively, providing high-fidelity reward signals without the need for additional pre-processing, thereby maintaining high training throughput.

\section{Additional Experimental Analyses}\label{sec:appendix_more_experiments}

\subsection{Evaluation on General VQA Benchmarks}\label{sec:appendix_general_vqa_benchmark}

While \methodname substantially improves the policy model on a diverse set of vision-to-code tasks, it is important to examine whether these gains come with any trade-offs on broader multimodal capabilities. In particular, reinforcement learning with task-specific rewards may induce distributional shifts or over-specialization, potentially degrading performance on general-purpose visual question answering (VQA) benchmarks.

To assess the generality of the learned policy, we evaluate the RL-trained model on a suite of established VQA benchmarks that are closely related to the structured visual domains considered in this work. Concretely, we select multiple chart- and document-centric benchmarks that require fine-grained visual perception, text recognition, and cross-modal reasoning, thereby serving as a diagnostic for whether \methodname-guided optimization preserves general visual understanding skills beyond code generation. These benchmarks include ChartQA~\cite{masry2022chartqa}, CharXiv\_DQ~\cite{wang2024charxiv}, CharXiv\_RQ~\cite{wang2024charxiv}, DocVQA~\cite{mathew2021docvqa} and InfoVQA~\cite{mathew2022infographicvqa}.

\begin{table*}[t]
\caption{\small \textbf{Results on general VQA benchmarks.} We evaluate the policy model trained with \methodname-guided RL on multiple general-purpose benchmarks (with an emphasis on chart- and document-centric VQA) to examine whether its improved vision-to-code performance comes at the cost of general multimodal capability.}
\vspace{+2pt}
\label{tab:general_benchmarks}
\centering
\setlength{\tabcolsep}{8pt}         
\renewcommand{\arraystretch}{.95}    
\resizebox{1.\linewidth}{!}{%
\begin{tabular}{@{} l | c c c c c | c}
\toprule[1.25pt]
\textbf{Models}
& \textbf{ChartQA\_TEST} & \textbf{CharXiv\_DQ} & \textbf{CharXiv\_RQ} & \textbf{DocVQA\_VAL} & \textbf{InfoVQA\_VAL} & \textbf{AVG} \\	
\midrule
Qwen3-VL-8B-Instruct  & 82.9 & 83.8 & 46.0 & 95.6 & 83.1 & 78.3 \\
\midrule
$+$ RL on Chart-to-Code  & 82.6 & 83.5 & 47.2 & 95.6 & 83.1 & 78.4 \\
$+$ RL on Table-to-Markdown  & 81.5 & 83.8 & 46.7 & 95.5 & 83.1 & 78.1 \\
$+$ RL on SVG-to-Code  & 83.2 & 83.5 & 46.4 & 95.8 & 83.6 & 78.5 \\
\bottomrule[1.25pt]
\end{tabular}%
}
\end{table*}

Tab.~\ref{tab:general_benchmarks} summarizes the results. Overall, \methodname-guided RL does not degrade general VQA performance: the average score remains essentially stable and even shows slight improvements for two of the three RL policies. Concretely, compared to the base Qwen3-VL-8B-Instruct (AVG 78.3), RL on Chart-to-Code reaches 78.4 (+0.1) and RL on SVG-to-Code reaches 78.5 (+0.2), while RL on Table-to-Markdown is comparable at 78.1 (-0.2). At the benchmark level, the most consistent change is on CharXiv-RQ, where all RL variants improve over the base (46.0 $\rightarrow$ 47.2 / 46.7 / 46.4), suggesting better fine-grained reasoning over chart-centric queries. Meanwhile, document-centric performance is preserved: DocVQA remains nearly unchanged (95.6 $\rightarrow$ 95.6 / 95.5 / 95.8), and InfoVQA is stable with a small gain for the SVG RL policy (83.1 $\rightarrow$ 83.6). These results indicate that the gains on vision-to-code do not arise from sacrificing general multimodal competence; instead, \methodname provides a largely non-destructive learning signal that improves structured visual parsing while maintaining broad chart/document understanding.

\subsection{Evaluation of the Effect of Multi-Task Data Mixing}\label{sec:appendix_data_mixing}

\subsubsection{Effect of Multi-Task Data Mixing on \benchname}\label{sec:data_mixing}
\methodname is trained to support multiple tasks that share partial failure modes but also exhibit task-specific characteristics. To examine whether joint training on mixed data is beneficial or induces negative transfer, we ablate the reward model’s training data composition by comparing a unified model trained on all three data sources with task-specific variants trained on a single data type. Results on \benchname are reported in Tab.~\ref{tab:data_mix_ablation}.

\begin{table*}[t]
\vspace{-4pt}
\caption{\small \textbf{Effect of Multi-Task Data Mixing.} We compare reward models trained on mixed multi-task data vs.\ single-task data.}
\vspace{+2pt}
\centering
\setlength{\tabcolsep}{10pt}
\renewcommand{\arraystretch}{.95}
\resizebox{1.\linewidth}{!}{%
\begin{tabular}{@{} l | *3{c} | *3{c} | *3{c} | *3{c} @{}}
\toprule[1.25pt]
\multirow{2}{*}{\textbf{Model}} 
& \multicolumn{3}{c}{\textbf{Chart}} & \multicolumn{3}{|c}{\textbf{Table}} & \multicolumn{3}{|c|}{\textbf{SVG}} & \multicolumn{2}{|c}{\textbf{AVG}} \\
\cmidrule(lr){2-4}\cmidrule(lr){5-7}\cmidrule(lr){8-10}\cmidrule(lr){11-13}
& $F1_h$ & $F1_s$ & $S_c$ & $F1_h$ & $F1_s$ & $S_c$ & $F1_h$ & $F1_s$ & $S_c$ & $F1_h$ & $F1_s$ & $S_c$ \\
\midrule
\addlinespace[2pt]
\rowcolor{gray!20}
\multicolumn{13}{c}{\textit{Visual-ERM}} \\
\midrule
$+$ Chart-Data-Only  & \cellcolor{gray!20} 40.1 & \cellcolor{gray!20} 42.6 & \cellcolor{gray!20} 58.1 & 17.2 & 20.4 & 10.5 & 11.3 & 15.3 & 50.1 & 26.3 & 29.3 & 23.7 \\
$+$ Table-Data-Only  & 17.1 & 20.1 & 49.3 & \cellcolor{gray!20} 53.6 & \cellcolor{gray!20} 54.9 & \cellcolor{gray!20} 68.8 & 14.1 & 19.0 & 55.2 & 31.7 & 34.2 & 56.6 \\
$+$ SVG-Data-Only  & 16.4 & 21.0 & 56.0 & 11.4 & 17.6 & 37.0 & \cellcolor{gray!20} 26.4 & \cellcolor{gray!20} 31.0 & \cellcolor{gray!20} 57.3 & 17.5 & 22.6 & 44.2 \\
\rowcolor[HTML]{DAEFF9} $+$ Mix-Data  & 39.9 & 42.8 & 61.2 & 56.4 & 57.6 & 74.8 & 28.3 & 32.6 & 59.6 & 42.1 & 44.7 & 58.4 \\
\bottomrule
\end{tabular}%
}

\label{tab:data_mix_ablation}
\vspace{-16pt}
\end{table*}

Mixing chart, table, and SVG data yields the best overall performance. We attribute this to cross-task transfer of error patterns (e.g., recognition and layout behaviors learned from tables can generalize to charts due to overlapping failure modes) leading to a mutually beneficial effect under joint training. In Sec.~\ref{sec:data_mixing_rl}, we further validate this finding by comparing the downstream RL performance of reward models trained on different data sources.


\begin{table*}[t]
\caption{\small \textbf{Ablation Study of Multi-Task Data Mixing}. We perform RL on the table parsing task using different reward models: one reward model is trained only on table data, while the other reward model is jointly trained on table, chart, and SVG data. TEDS-S denotes TEDS-Structure-Only, and TA represents the table subtask of olmOCRBench. For Edit-Dist, where lower values indicate better performance, we use ($100-\text{Edit-Dist}$) when computing the Average.}
\vspace{+2pt}
\label{tab:data_mixing}
\centering
\setlength{\tabcolsep}{8pt}         
\renewcommand{\arraystretch}{1.}    
\resizebox{1.\linewidth}{!}{%
\begin{tabular}{@{} l *3{c} *3{c} c c @{}}
\toprule[1.25pt]
\multirow{2}{*}{\textbf{Model}} 
& \multicolumn{3}{c}{\textbf{OmniDocBench}} & \multicolumn{3}{c}{\textbf{OmniDocBench-v1.5}} & \multicolumn{1}{c}{\textbf{olmOCRBench}} & \multirow{2}{*}{\textbf{Avg}$\uparrow$} \\
\cmidrule(lr){2-4}\cmidrule(lr){5-7}\cmidrule(lr){8-8}
& TEDS$\uparrow$ & TEDS-S$\uparrow$ & Edit-Dist$\downarrow$ & TEDS$\uparrow$ & TEDS-S$\uparrow$ & Edit-Dist$\downarrow$ & TA$\uparrow$ & \\
\midrule
\rowcolor{gray!20}
\multicolumn{9}{c}{\textit{Baseline}} \\
\midrule
Qwen3-VL-8B-Instruct  & 78.9 & 83.9 & 23.2 & 72.7 & 77.2 & 26.9 & 75.3 & 76.8\\
\midrule
\rowcolor{gray!20}
\multicolumn{9}{c}{\textit{RL with Different Reward Model}} \\
\midrule
\rowcolor[HTML]{DAEFF9} + RM (table-data-only)  & \underline{79.5} & \underline{84.4} & \underline{21.6} & \underline{74.4} & \underline{79.4} & \underline{25.0} & \textbf{79.4} & \underline{78.6} \\
$\Delta$  & \hgreen{+0.6} & \hgreen{+0.5} & \hgreen{+1.6} & \hgreen{+1.7} & \hgreen{+2.2} & \hgreen{+1.9} & \hgreen{4.1} & \hgreen{+1.8} \\
\midrule
\rowcolor[HTML]{DAEFF9} + RM (mix-data,\methodname)  & \textbf{81.4} & \textbf{86.3} & \textbf{20.7} & \textbf{75.4} & \textbf{80.4} & \textbf{24.2} & \underline{78.1} & \textbf{79.5} \\
$\Delta$  & \hgreen{+2.5} & \hgreen{+2.4} & \hgreen{+2.5} & \hgreen{+2.7} & \hgreen{+3.2}  & \hgreen{+2.7} & \hgreen{+2.8} & \hgreen{+2.7} \\
\addlinespace[2pt]
\bottomrule[1.25pt]
\end{tabular}%
}
\end{table*}

As shown in Tab.~\ref{tab:data_mix_ablation}, reward models trained on single-task data exhibit a strong task bias. For example, Chart-Data-Only performs well on the Chart subtask ($F1_h=40.1, F1_s=42.6$), but performs much worse on Table and SVG tasks (e.g., Table $F1_h=17.2$, SVG $F1_h=11.3$). Similarly, although Table-Data-Only achieves strong results on the Table subtask ($F1_h=53.6$, $S_c=68.8$), it generalizes poorly to Chart ($F1_h=17.1$). In contrast, the mixed-data RM (Mix-Data) achieves more balanced and consistently strong performance across all three task categories. It maintains nearly the same performance on Chart ($F1_h=39.9$ vs. 40.1), while significantly improving on Table and SVG (Table: $F1_h=56.4$, SVG: $F1_h=28.3$). As a result, it attains the best average performance (AVG: $F1_h=42.1, F1_s=44.7, S_c=58.4$). These results indicate that multi-task training does not introduce obvious negative transfer; instead, it provides positive transfer by sharing structured understanding and code generation abilities, leading to a more generalizable reward model.

\subsubsection{Effect of Multi-Task Data Mixing on RL}\label{sec:data_mixing_rl}
To further validate this phenomenon in a practical setting, we apply RMs trained with different datasets to actual RL training and evaluate the effects in real usage. We select the Table-to-Markdown task for this experiment, and the results are shown in Tab.~\ref{tab:data_mixing}. As shown in Tab.~\ref{tab:data_mixing}, both the Table-only RM and the mixed-data RM lead to stable improvements in the Table-to-Markdown RL scenario. Compared with the baseline, RM (table-data-only) improves TEDS and TEDS-S on both benchmarks and reduces Edit-Dist, resulting in an overall average gain of +1.8. More importantly, the mixed-data RM achieves further improvements over the single-task RM on most metrics. On OmniDocBench, it reaches $TEDS=81.4$ and $TEDS\text{-}S=86.3$, while further reducing Edit-Dist to 20.7. On olmOCRBench-v1.5, it also achieves higher TEDS/TEDS-S and lower Edit-Dist. Overall, it yields an average improvement of +2.7. Although the mixed RM shows a slight drop on the TA metric compared to the Table-only RM (78.1 vs. 79.4), the overall gains are larger, suggesting that a richer multi-task reward signal provides a more robust optimization direction for RL, improving both final performance and generalization on real Table-to-Markdown tasks.

\subsection{Ablation on Different Judge Models}\label{sec:appendix_judge_model_ablation}

\benchname evaluates a reward model’s ability to identify fine-grained image-to-image discrepancies and vision-to-code reconstruction errors. In addition to objective metrics (e.g., scalar scores, types and severities), \benchname also includes \emph{free-form} discrepancy descriptions. Since these descriptions are open-ended and may use different wording to refer to the same underlying issue, an exact string match is insufficient for reliable evaluation.

\paragraph{LLM-assisted matching protocol.}
To evaluate such free-form outputs in a scalable and consistent manner, we employ an LLM-as-Judge to perform \emph{error-level alignment} between the reward model’s predicted error list and the ground-truth error list. Specifically, given the predicted errors and the annotated errors, the judge determines for each predicted item whether it (i) correctly matches a ground-truth error, (ii) is a hallucinated/unsupported claim, or (iii) corresponds to an error that exists but is described incorrectly. Symmetrically, the judge also identifies ground-truth errors that are missed by the model. Based on this alignment, we compute two variants of F1:
(i) a strict matching criterion ($F1_h$, \emph{hard}) that requires close semantic and attribute-level agreement, and
(ii) a relaxed criterion ($F1_s$, \emph{soft}) that tolerates minor paraphrases or partial matches, reflecting a more lenient notion of equivalence.

\paragraph{Why judge ablations matter.}
Using an LLM introduces a potential dependency on the judge’s calibration and reasoning style. However, in \benchname the judge is only asked to perform relatively constrained \emph{matching} (rather than open-ended preference ranking), which we hypothesize to be less sensitive to the specific judge model. To validate this, we rerun the same \methodname predictions on \benchname while varying the judge among several widely-used API models, and report the resulting scores in Tab.~\ref{tab:llm_as_judge}.

\begin{table*}[t]
\caption{\small \textbf{Ablation on LLM Judges.} \benchname uses LLM-assisted judging to verify answer correctness. To assess the robustness of this evaluation protocol, we compare multiple LLM-as-Judge models and report the resulting consistency across judges.}
\vspace{+2pt}
\label{tab:llm_as_judge}
\centering
\setlength{\tabcolsep}{8pt}         
\renewcommand{\arraystretch}{.95}    
\resizebox{1.\linewidth}{!}{%
\begin{tabular}{@{} l | *3{c} | *3{c} | *3{c} | *3{c} @{}}
\toprule[1.25pt]
\multirow{2}{*}{\textbf{Judger}} 
& \multicolumn{3}{c}{\textbf{Chart}} & \multicolumn{3}{|c}{\textbf{Table}} & \multicolumn{3}{|c|}{\textbf{SVG}} & \multicolumn{2}{|c}{\textbf{AVG}} \\
\cmidrule(lr){2-4}\cmidrule(lr){5-7}\cmidrule(lr){8-10}\cmidrule(lr){11-13}
& $F1_h$ & $F1_s$ & $S_c$ & $F1_h$ & $F1_s$ & $S_c$ & $F1_h$ & $F1_s$ & $S_c$ & $F1_h$ & $F1_s$ & $S_c$ \\
\midrule
GPT-5-mini  & 39.9 & 42.8 & 61.2 & 56.4 & 57.6 & 74.8 & 28.3 & 32.6 & 59.6 & 42.1 & 44.7 & 58.4 \\
GPT-5.2  & 40.9 & 45.4 & 61.2 & 55.0 & 57.1 & 74.8 & 24.5 & 31.6 & 59.6 & 40.8 & 45.1 & 58.4 \\
Gemini2.5-Pro  & 40.8 & 42.1 & 61.2 & 54.8 & 55.7 & 74.8 & 27.1 & 28.7 & 59.6 & 41.2 & 42.4 & 58.4 \\
Gemini3-Flash  & 41.5 & 43.7 & 61.2 & 56.5 & 57.8 & 74.8 & 28.5 & 32.1 & 59.6 & 43.0 & 44.9 & 58.4 \\
\bottomrule[1.25pt]
\end{tabular}%
}
\end{table*}

\paragraph{Results and analysis.}
Tab.~\ref{tab:llm_as_judge} shows that the evaluation is highly stable across judges.
First, the $S_c$ is identical for all judges (e.g., Chart: 61.2, Table: 74.8, SVG: 59.6, AVG: 58.4), as expected since it is computed deterministically and does not rely on the LLM matcher.
Second, the judge-dependent metrics ($F1_h$ and $F1_s$) vary only modestly: on average, $F1_h$ ranges from 40.8 to 43.0 (a span of 2.2 points) and $F1_s$ ranges from 42.4 to 45.1 (a span of 2.7 points) across all judges. At the task level, the variability remains limited—for instance, Table $F1_s$ ranges from 55.7 to 57.8—indicating that different judges largely agree on which predicted errors are correct versus hallucinated or missing.

Overall, these results suggest that \benchname is not overly sensitive to the choice of LLM judge. Since the judge is used for structured error-to-error matching under well-specified criteria, replacing the judge model yields consistent conclusions, supporting the robustness of our evaluation protocol.

\subsection{Ablation Study on Reward Design}\label{sec:appendix_reward_design}
We investigate two reward designs. The first directly uses the scalar score produced by the reward model as the reward signal. To improve training stability, the second design augments the reward with an additional render-success reward that indicates whether the generated output can be successfully rendered, which is similar to a format reward. To combine the render-success reward with the reward-model score in a principled manner, we further apply score normalization before aggregation.

\begin{table*}[t]
\caption{\small \textbf{Ablation Study of Reward Design}. \textit{RSR} denotes the Render-Success Reward. TEDS-S denotes TEDS-Structure-Only, and TA represents the table subtask of olmOCRBench. For Edit-Dist, where lower values indicate better performance, we use ($100 - \text{Edit-Dist}$) when computing the Average.}
\vspace{+2pt}
\label{tab:reward_design_ablation}
\centering
\setlength{\tabcolsep}{4pt}         
\renewcommand{\arraystretch}{1.}    
\resizebox{1.\linewidth}{!}{%
\begin{tabular}{@{} l *3{c} *3{c} c c @{}}
\toprule[1.25pt]
\multirow{2}{*}{\textbf{Model}} 
& \multicolumn{3}{c}{\textbf{OmniDocBench}} & \multicolumn{3}{c}{\textbf{OmniDocBench-v1.5}} & \multicolumn{1}{c}{\textbf{olmOCRBench}} & \multirow{2}{*}{\textbf{Avg}$\uparrow$} \\
\cmidrule(lr){2-4}\cmidrule(lr){5-7}\cmidrule(lr){8-8}
& TEDS$\uparrow$ & TEDS-S$\uparrow$ & Edit-Dist$\downarrow$ & TEDS$\uparrow$ & TEDS-S$\uparrow$ & Edit-Dist$\downarrow$ & TA$\uparrow$ & \\
\midrule
Qwen3-VL-8B-Instruct  & 78.9 & 83.9 & 23.2 & 72.7 & 77.2 & 26.9 & 75.3 &76.8\\
GRPO (w/o RSR)  & 80.9 & 85.6 & 21.8 & 75.2 & 79.9 & 25.9 & 79.0 & 79.0 \\
\rowcolor[HTML]{DAEFF9} GRPO (w RSR) & 81.4 & 86.3 & 20.7 & 75.4 & 80.4 & 24.2 & 78.1 & 79.5 \\
\addlinespace[2pt]
\bottomrule[1.25pt]
\end{tabular}%
}
\end{table*}

\begin{figure*}[http]
    \begin{center}
    \includegraphics[width=1.\linewidth]{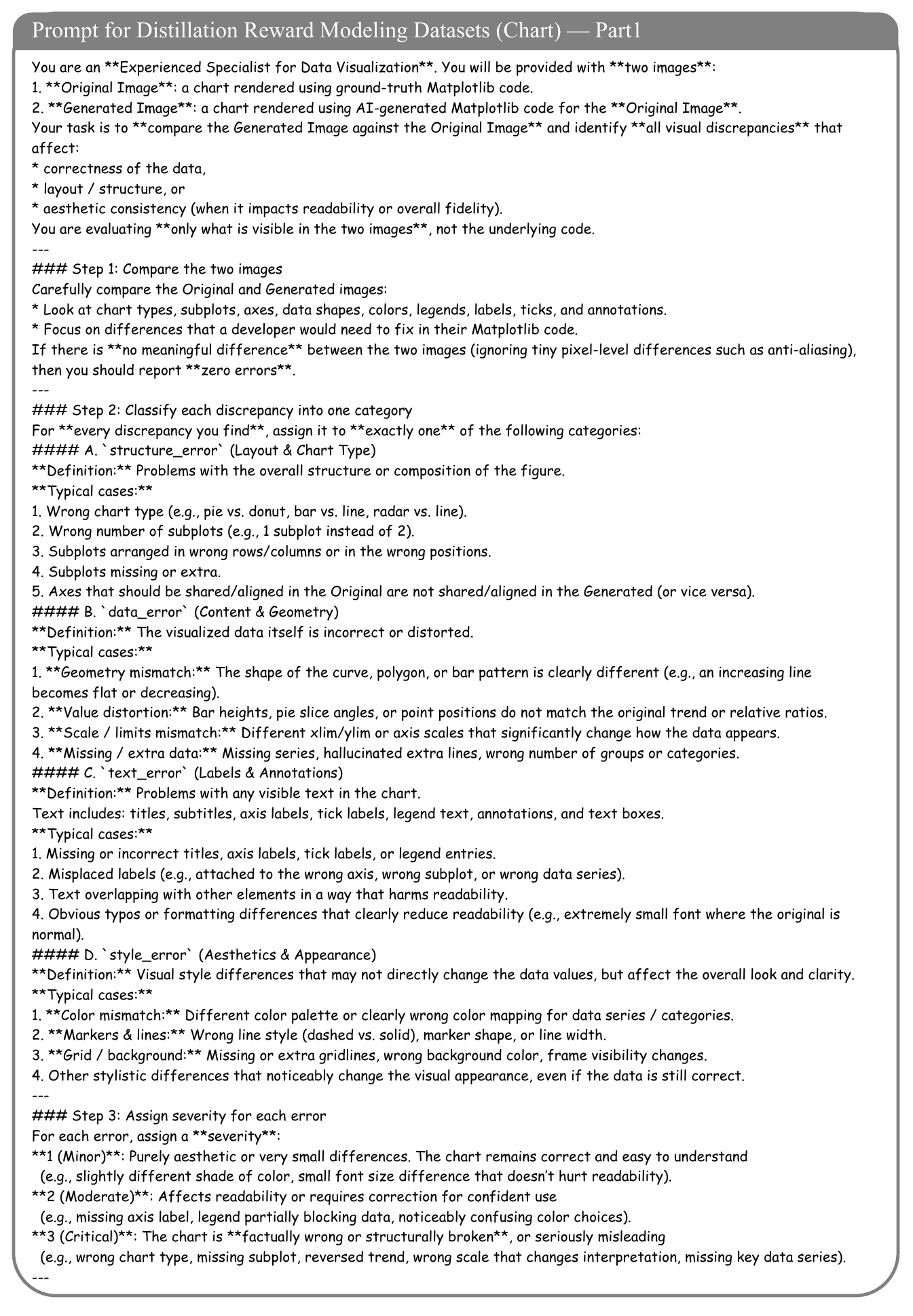}
    \end{center}
    \vspace{-12pt}
    \caption{\small \textbf{Prompt for distillation.} The prompt used to distill GPT-5-mini to construct reward-modeling training data for \methodname. We show the Chart-specific prompt here; due to its length, this table includes only the first part.}
    \label{fig:Distillation_pt1}
\end{figure*}

\begin{figure*}[h]
    \begin{center}
    \includegraphics[width=1.\linewidth]{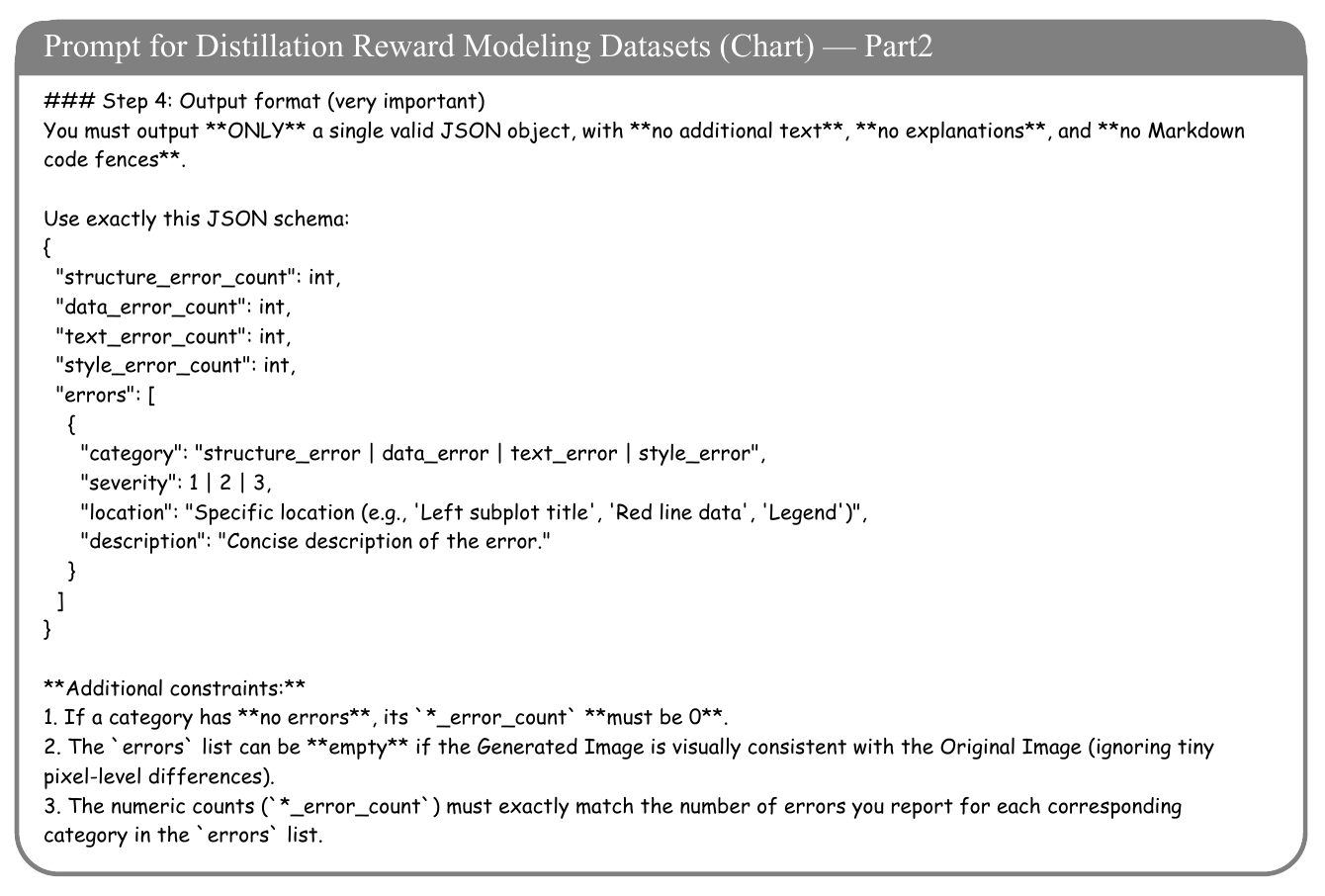}
    \end{center}
    \vspace{-12pt}
    \caption{\small \textbf{Prompt for distillation.} The prompt used to distill GPT-5-mini to construct reward-modeling training data for \methodname. We show the Chart-specific prompt here; due to its length, this table includes only the second part.}
    \label{fig:Distillation_pt2}
\end{figure*}

We conduct an ablation study comparing these two reward formulations and report the results in Tab.~\ref{tab:reward_design_ablation}. While both designs yield comparable performance gains, the formulation incorporating a render-success reward achieves a slightly larger improvement, which we attribute to its enhanced training stability. Specifically, we observe that during the initial stages of training, the model occasionally produces code that cannot be successfully rendered, leading to intermittent zero rewards. However, the policy rapidly learns to navigate these formatting constraints, allowing the optimization focus to transition quickly from basic renderability to the primary objective of preserving fine-grained visual characteristics. This phased learning trajectory ensures that the model effectively prioritizes structural and semantic fidelity once the fundamental format requirements are mastered.

\subsection{More Test-Time Scaling Results}
\label{sec:appendix_tts}

In the main paper, we evaluate reflection-based test-time scaling (TTS) on Chart-to-Code. To further validate the effectiveness and generality of this inference-time refinement strategy, we additionally conduct TTS experiments on SVG-to-Code, with results reported in Tab.~\ref{tab:svg_reflection}. We report SSIM and CLIPScore (higher is better) and LPIPS (lower is better).

The table contains two independent comparisons. 
The first compares Qwen3-VL-8B-Instruct with and without \methodname-guided reflection. Reflection yields consistent improvements across all metrics: CLIPScore increases from 73.3 to 74.6, SSIM remains comparable (60.9 vs.\ 60.4), and LPIPS decreases from 60.0 to 58.1 (improved). Overall, the aggregated score improves from 64.2 to 65.2 (+1.0), demonstrating that \methodname can provide actionable feedback for inference-time refinement even without RL.

The second comparison evaluates TTS on the \methodname-guided RL policy (Qwen3-VL-8B-Instruct + RL on \methodname). In this setting, applying reflection further boosts performance: LPIPS stays essentially stable (49.7 vs.\ 49.2, improved), and CLIPScore increases from 77.5 to 78.1. The overall score rises from 69.4 to 69.8 (+0.4). These results indicate that reflection provides additional gains on top of RL, and the RL-trained policy benefits from a stronger starting point, achieving the best absolute performance in both CLIPScore and overall score.


\begin{table}[t]
\vspace{-4pt}
\centering

\begin{minipage}[t]{0.62\linewidth}
\vspace{0pt} 
\centering
\setlength{\tabcolsep}{2pt}
\renewcommand{\arraystretch}{.90}
\resizebox{\linewidth}{!}{%
\begin{tabular}{@{} l *4{c} @{}}
\toprule[1.25pt]
\multirow{2}{*}{Model} 
& \multicolumn{4}{c}{\textbf{UniSVG(ISVGEN)}} \\
\cmidrule(lr){2-5}
& SSIM$\uparrow$ & LPIPS$\downarrow$ & CLIP$\uparrow$ & \textbf{Score}$\uparrow$ \\
\midrule
Qwen3-VL-8B-Instruct  & 60.9& 60.0& 73.3& 64.2   \\ 
\rowcolor[HTML]{DAEFF9} 
$+$ Reflection (Visual-ERM) & 60.4 & 58.1 & 74.6 & 65.2 \\
\midrule
Qwen3-VL-8B-Instruct (+RL on Visual-ERM)   & 64.3& 49.7& 77.5& 69.4   \\ 
\rowcolor[HTML]{DAEFF9} 
$+$ Reflection (Visual-ERM) & 63.9 & 49.2 & 78.1 & 69.8 \\
\bottomrule[1.25pt]
\end{tabular}%
}
\end{minipage}
\hfill
\begin{minipage}[t]{0.35\linewidth}
\vspace{0pt} 
\caption{\small \textbf{Test-Time Scaling on SVG-to-Code}. We evaluate TTS on the \textit{SVG-to-Code} task. Leveraging \methodname as an evaluator, the model performs multiple rounds of self-reflection and revision, leading to improved parsing performance.}
\label{tab:svg_reflection}
\end{minipage}

\end{table}

\subsection{Scalability of \methodname}

\paragraph{Discrepancy Patterns.} The primary categories defined in our training are macro-level abstractions that encompass nearly the entire spectrum of errors in visual-to-code tasks. We observe that almost all fine-grained discrepancies, including those not explicitly seen during training, are effectively combinations or specific instances of these core failure modes. By grounding the reward model in these fundamental dimensions, \methodname develops a generalized capability to identify and penalize diverse error patterns regardless of their novelty.

Our data generation strategy further reinforces this adaptability through two synergistic modes:
\textbf{1) Systematic Coverage (Targeted Edit):} We utilize high-capacity models to deliberately perturb structured representations, ensuring that the reward model maps the complete theoretical space of structural and semantic failures.
\textbf{2) Empirical Realism (Natural Inference):} We sample authentic errors directly from weaker models to align the training distribution with the messy, unpredictable failure modes encountered in practice.

Together, these modes ensure that \methodname does not merely memorize a fixed list of labels but instead masters the underlying logic of visual-to-code equivalence. This allows the model to remain robust and precise when evaluating complex, hybrid, or previously unencountered discrepancies.

\paragraph{Broadening Task Scope.} While our current evaluation focuses on structured visuals such as charts, tables, and SVGs, \methodname is designed to capture \textbf{atomic discrepancy patterns} that are fundamentally shared across nearly all vision-to-code domains. We argue that the core challenges in tasks like UI reconstruction or mathematical diagramming are not unique, but are instead composed of the same structural and semantic building blocks mastered by our model. For instance, the spatial layout and color fidelity critical for UI reconstruction are directly mirrored in our evaluation of complex chart and table structures.

In essence, \methodname does not just learn task-specific rules; it learns the universal rules of visual-to-code equivalence. Because these "unseen" tasks rely on the same underlying principles of preserving shape, color, and structure, \methodname can be naturally extended to a wider range of vision-to-code scenarios without requiring fundamental architectural changes.

\section{Prompts}\label{sec:appendix_prompts}

\methodname relies on several designed prompting templates at multiple stages of the pipeline. Specifically, we use prompts for: (i) distilling GPT-5-mini to construct reward-modeling training data for \methodname, (ii) \methodname inference across different vision-to-code tasks (Chart-to-Code, Table-to-Markdown, and SVG-to-Code), and (iii) the LLM-assisted judging protocol used in \benchname for evaluating subjective, free-form error descriptions. We provide the full prompts in this section for reproducibility.

\subsection{Distillation Prompt}\label{sec:appendix_distillation_prompt}

To train \methodname, we first build a reward-modeling dataset by distilling GPT-5-mini. Each training instance consists of an \emph{image pair}: an \textbf{Original Image} and a \textbf{Generated Image} rendered from a candidate output (e.g., model-generated code/markdown). The distillation target is a structured comparison: the model must identify and describe \emph{all visually meaningful discrepancies} between the two images, focusing strictly on what is observable in the rendered results rather than the underlying code.

We construct these image pairs via an automated process that combines (i) \emph{error injection} using a stronger yet cost-effective model (GPT-5-mini) to synthesize diverse, realistic failure patterns, and (ii) \emph{lightweight policy inference} to mimic errors produced by smaller models in practice. The resulting pairs, together with a detailed instruction prompt, are provided to GPT-5-mini to generate high-quality supervision signals for reward modeling. Figures~\ref{fig:Distillation_pt1} and \ref{fig:Distillation_pt2} show the Chart-specific distillation prompt (split into two parts due to length).

\paragraph{Prompt overview (Chart).}
The prompt casts the judge as an ``experienced specialist for data visualization'' and explicitly defines the evaluation objective: compare the Generated Image against the Original Image and report discrepancies that affect (i) correctness of the data, (ii) layout/structure fidelity, and (iii) visual consistency/readability. Importantly, it instructs the judge to output \texttt{zero errors} if no meaningful difference exists.

\paragraph{Error taxonomy and severity.}
To ensure consistent and fine-grained supervision, the prompt requires every discovered discrepancy to be assigned to exactly one of four categories:
\emph{structure\_error} (layout and chart type), \emph{data\_error} (content/geometry and value distortion), \emph{text\_error} (titles, labels, legends, ticks, and annotations), and \emph{style\_error} (aesthetic and appearance). For each error, the judge also assigns a severity level on a three-point scale (minor / moderate / critical), which captures how strongly the discrepancy impacts fidelity or interpretation. This taxonomy is designed to reflect common failure modes in chart rendering and to provide training targets that are both interpretable and actionable.

\paragraph{Structured JSON output.}
Beyond natural-language descriptions, the prompt enforces a strict output format. As shown in Fig.~\ref{fig:Distillation_pt2}, the judge must output \textbf{only} a single valid JSON object following a fixed schema, including: (i) per-category error counts, and (ii) an explicit list of error items, each annotated with \texttt{category}, \texttt{severity}, a localized \texttt{location} description (e.g., ``legend'', ``left subplot title''), and a concise \texttt{description}. The prompt additionally specifies consistency constraints (e.g., counts must match the number of listed errors; the error list must be empty when the two images are visually consistent). These constraints substantially reduce annotation ambiguity and make the distilled outputs directly usable for reward-model training and evaluation.

\paragraph{Why this design.}
This prompt design provides two practical benefits. First, it produces \emph{fine-grained} supervision that decomposes visual discrepancies into interpretable components (structure/data/text/style) rather than collapsing them into a single opaque score. Second, the enforced JSON format enables scalable filtering, aggregation, and auditing of the distilled labels, which is critical for building large reward-modeling datasets with reliable quality control.

\begin{figure*}[t]
    \begin{center}
    \includegraphics[width=1.\linewidth]{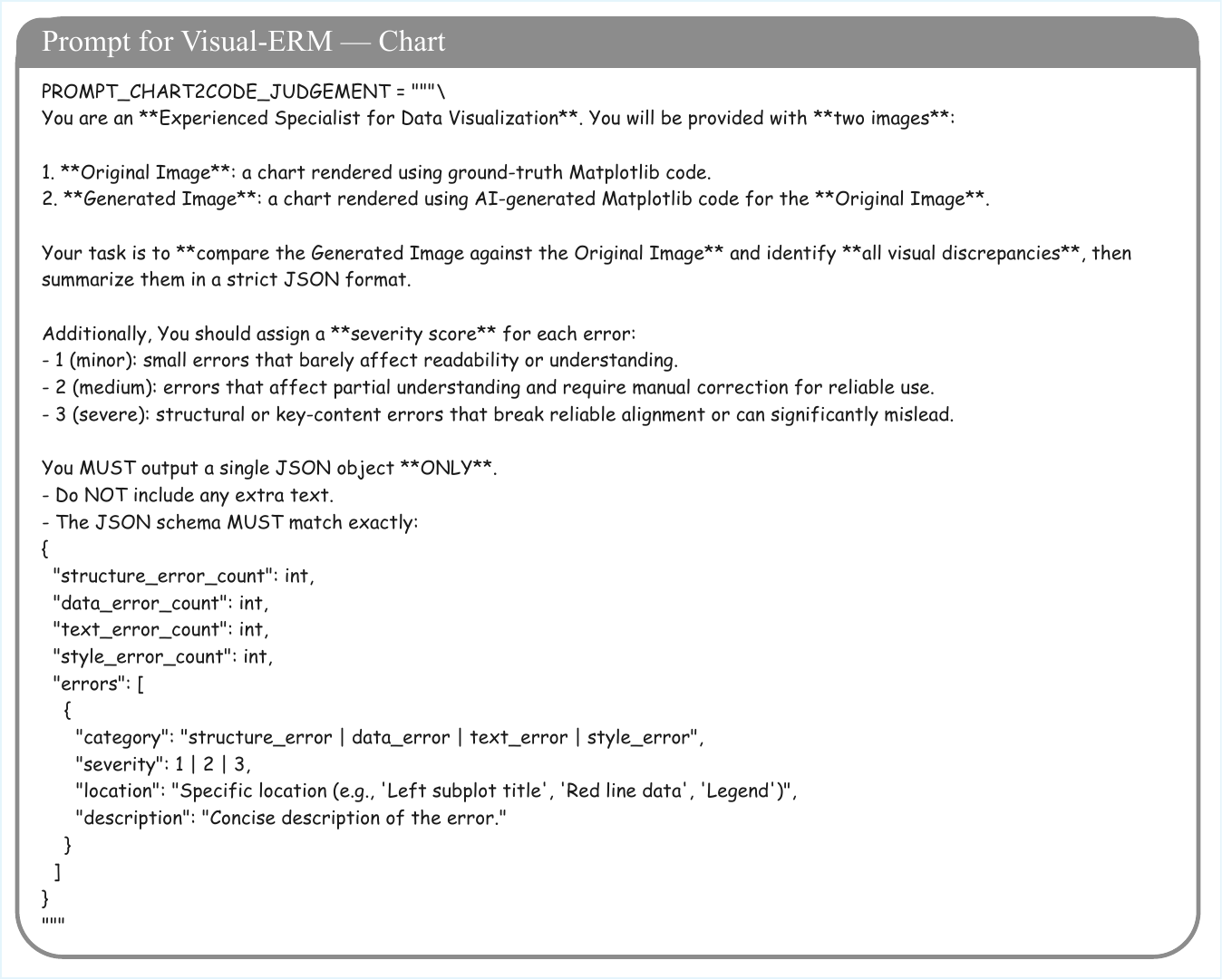}
    \end{center}
    \vspace{-12pt}
    \caption{\small \textbf{\methodname inference prompt.} The prompting template used by \methodname at inference time. This prompt is specialized for the \textbf{Chart-to-Code} setting.}
    \label{fig:RM_pt_Chart}
\end{figure*}

\begin{figure*}[t]
    \begin{center}
    \includegraphics[width=1.\linewidth]{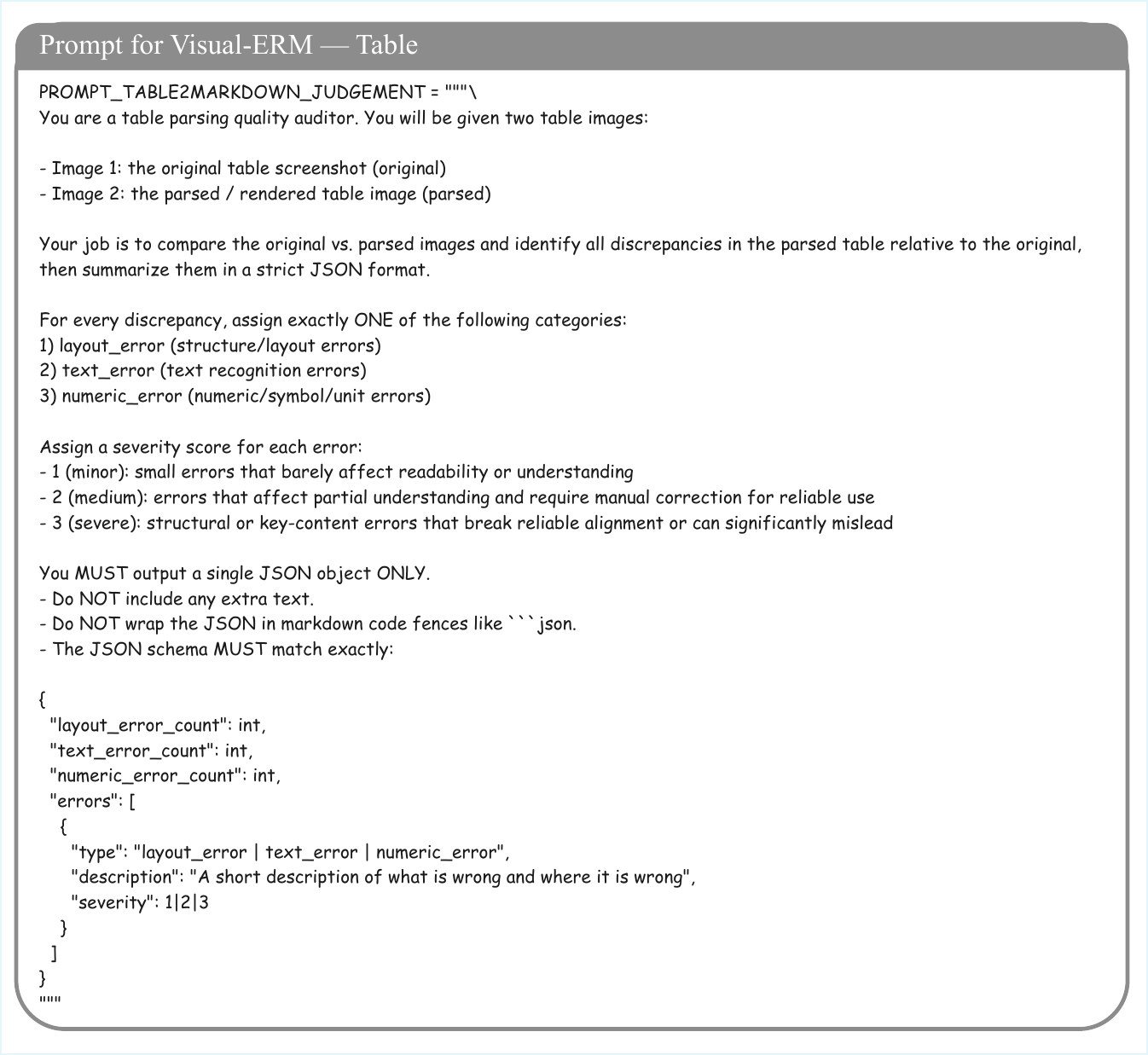}
    \end{center}
    \vspace{-12pt}
    \caption{\small \textbf{\methodname inference prompt.} The prompting template used by \methodname at inference time. This prompt is specialized for the \textbf{Table-to-Markdown} setting.}
    \label{fig:RM_pt_Table}
\end{figure*}

\begin{figure*}[t]
    \begin{center}
    \includegraphics[width=1.\linewidth]{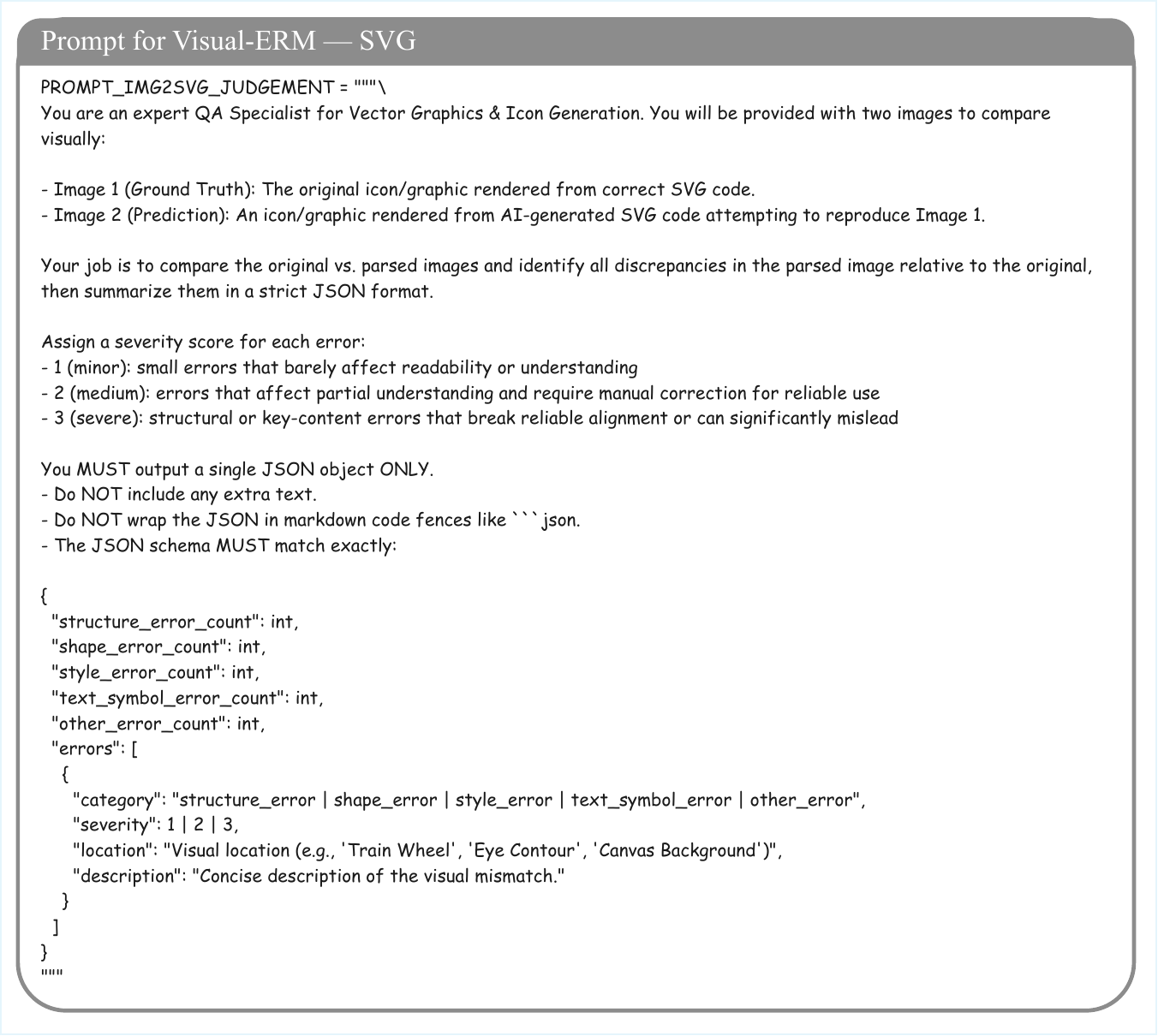}
    \end{center}
    \vspace{-12pt}
    \caption{\small \textbf{\methodname inference prompt.} The prompting template used by \methodname at inference time. This prompt is specialized for the \textbf{SVG-to-Code} setting.}
    \label{fig:RM_pt_SVG}
\end{figure*}

\subsection{\methodname Inference Prompts}\label{sec:appendix_rm_inference_prompts}

After constructing the distillation dataset, we train our reward model, \methodname, to produce fine-grained, structured feedback for image-to-image visual equivalence. Since this behavior is learned through supervised fine-tuning, \methodname does not require an overly restrictive prompt at inference time. Instead, we adopt lightweight and efficient prompting templates that primarily (i) specify the comparison setup (Original vs.\ Generated), (ii) request exhaustive discrepancy discovery conditioned on the task, and (iii) enforce a strict JSON output schema.

We design a task-specific prompt for each vision-to-code setting—Chart-to-Code, Table-to-Markdown, and SVG-to-Code. These prompts share the same overall structure and interface, but differ in the error taxonomy and schema fields to reflect domain-specific failure modes (e.g., chart type/axes/legend for charts, layout and numeric/text errors for tables, and shape/style/text-symbol discrepancies for SVGs). The complete prompts are provided in Fig.~\ref{fig:RM_pt_Chart}, Fig.~\ref{fig:RM_pt_Table}, and Fig.~\ref{fig:RM_pt_SVG}.

\paragraph{Chart-to-Code prompt (Fig.~\ref{fig:RM_pt_Chart}).}
The chart prompt frames \methodname as a data visualization specialist and focuses the comparison on discrepancies that matter for faithful chart reproduction. It explicitly decomposes errors into \texttt{structure\_error}, \texttt{data\_error}, \texttt{text\_error}, and \texttt{style\_error}, covering (i) global composition (chart type, subplot arrangement, axes sharing), (ii) geometric/value fidelity (bar heights, line trends, point positions, scale mismatches), (iii) textual correctness (titles, tick labels, legends, annotations), and (iv) appearance-level mismatches that may affect readability. This taxonomy is particularly suitable for charts because visually minor differences (e.g., small font changes) can be less important than semantic distortions (e.g., wrong scale or swapped series), and the explicit severity field further encourages \methodname to surface high-impact errors prominently. In practice, the chart prompt also encourages localized descriptions (e.g., ``legend'', ``x-axis ticks'', ``left subplot line'') which makes the feedback more actionable for downstream debugging and revision (such as test-time scaling).

\paragraph{Table-to-Markdown prompt (Fig.~\ref{fig:RM_pt_Table}).}
The table prompt casts \methodname as a table parsing quality auditor and emphasizes structural alignment and textual/numeric accuracy. Compared to charts, tables have less ambiguity in global structure but are highly sensitive to \emph{layout} (row/column spans, merged cells, alignment) and \emph{cell-level content}. Accordingly, the prompt uses a concise three-way categorization: \texttt{layout\_error}, \texttt{text\_error}, and \texttt{numeric\_error}. This design reflects typical failure patterns in table parsing: slight layout shifts can cascade into many downstream cell mismatches, while OCR-style errors often manifest as character substitutions or missing tokens. Separating \texttt{numeric\_error} from \texttt{text\_error} is also important in practice, since numerical discrepancies (units, decimal points, sign errors) are often more harmful even when visually subtle. The enforced JSON schema provides a predictable interface for aggregating errors and computing metrics, while keeping the prompt lightweight enough for efficient inference.

\paragraph{SVG-to-Code prompt (Fig.~\ref{fig:RM_pt_SVG}).}
The SVG prompt targets vector-graphic reconstruction, where fidelity hinges on geometric primitives and styling attributes. To capture these characteristics, the prompt introduces categories beyond those used for raster-like structured visuals: \texttt{shape\_error} (incorrect geometry or missing/extraneous components), \texttt{style\_error} (stroke width, fill, color, opacity), \texttt{text\_symbol\_error} (glyphs, labels, icons) and \texttt{structure\_error} (global composition and grouping). This taxonomy reflects the fact that SVG failures often involve precise contours and layering, which may not be well described by ``data/text/style'' alone. Requiring a location field further encourages spatial grounding, making the output interpretable and enabling targeted correction during RL or test-time scaling.

\begin{figure*}[http]
    \begin{center}
    \includegraphics[width=1.\linewidth]{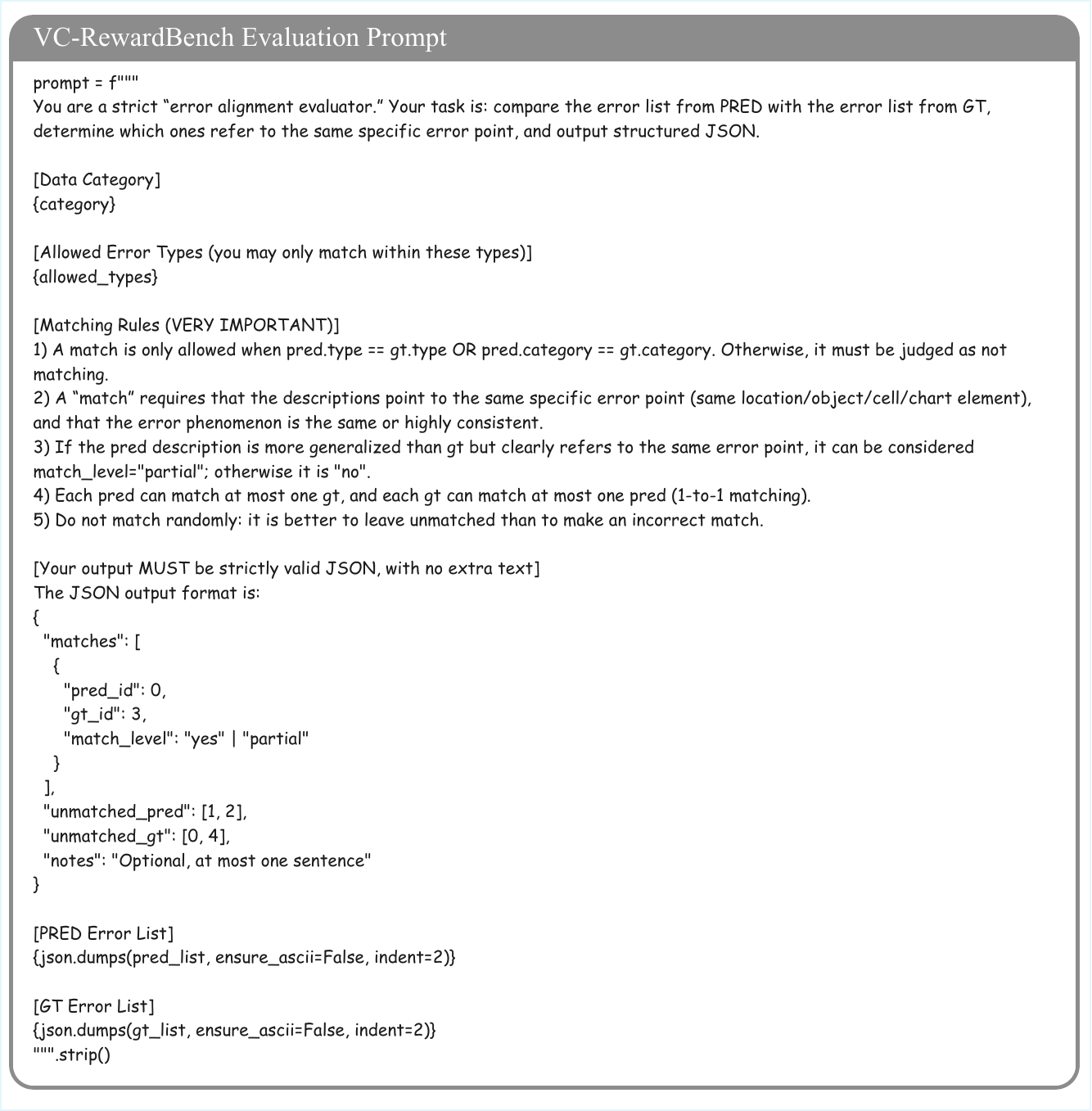}
    \end{center}
    \vspace{-12pt}
    \caption{\small \textbf{\benchname evaluation prompt.} For the subjective, free-form error descriptions in \benchname, we use an LLM-assisted judge to perform error matching and verification; the prompt used in this evaluation protocol is shown here.}
    \label{fig:bench_pt}
\end{figure*}

\paragraph{Common design choices across tasks.}
Across all three prompts, we adopt several shared choices that improve usability and robustness: (i) the judge is instructed to compare \emph{only what is visible} in the rendered images, preventing reliance on spurious code cues; (ii) the output must be a \emph{single valid JSON} object with a fixed schema, which reduces post-processing ambiguity and supports scalable training/evaluation; and (iii) the severity score provides a unified notion of error impact, enabling downstream pipelines (e.g., reflection-and-revision) to prioritize the most critical issues first. Together, these prompts serve as a lightweight interface that activates \methodname's learned capabilities while remaining consistent across heterogeneous vision-to-code domains.

\subsection{\benchname Evaluation Prompt}\label{sec:appendix_bench_eval_prompt}

A key component of \benchname is the evaluation of \emph{free-form error descriptions}, which summarize the visual mismatches between the original image and the image re-rendered from the parsed output. Since these descriptions are subjective and highly open-ended, rule-based string matching is brittle and fails to account for paraphrases, partial matches, or differences in granularity. We therefore use an LLM-as-Judge to perform verification: it determines which predicted errors are correctly matched to ground-truth errors, which are hallucinated, and which are only partially aligned.

Figure~\ref{fig:bench_pt} shows the prompt used for this evaluation. The prompt is intentionally designed to be \emph{constrained} and \emph{deterministic}: it restricts matching to a predefined set of error types, enforces strict type/category consistency, and requires 1-to-1 alignment between predicted and ground-truth error items. It also explicitly encourages conservative decisions (preferring ``unmatched'' over uncertain matches) and outputs a single JSON object with match indices and match levels (\texttt{yes} vs.\ \texttt{partial}). These constraints reduce judge arbitrariness and make the resulting metrics more robust across different judge models (as further validated in Tab.~\ref{tab:llm_as_judge}).

\begin{figure*}[http]
    \begin{center}
    \includegraphics[width=1.\linewidth]{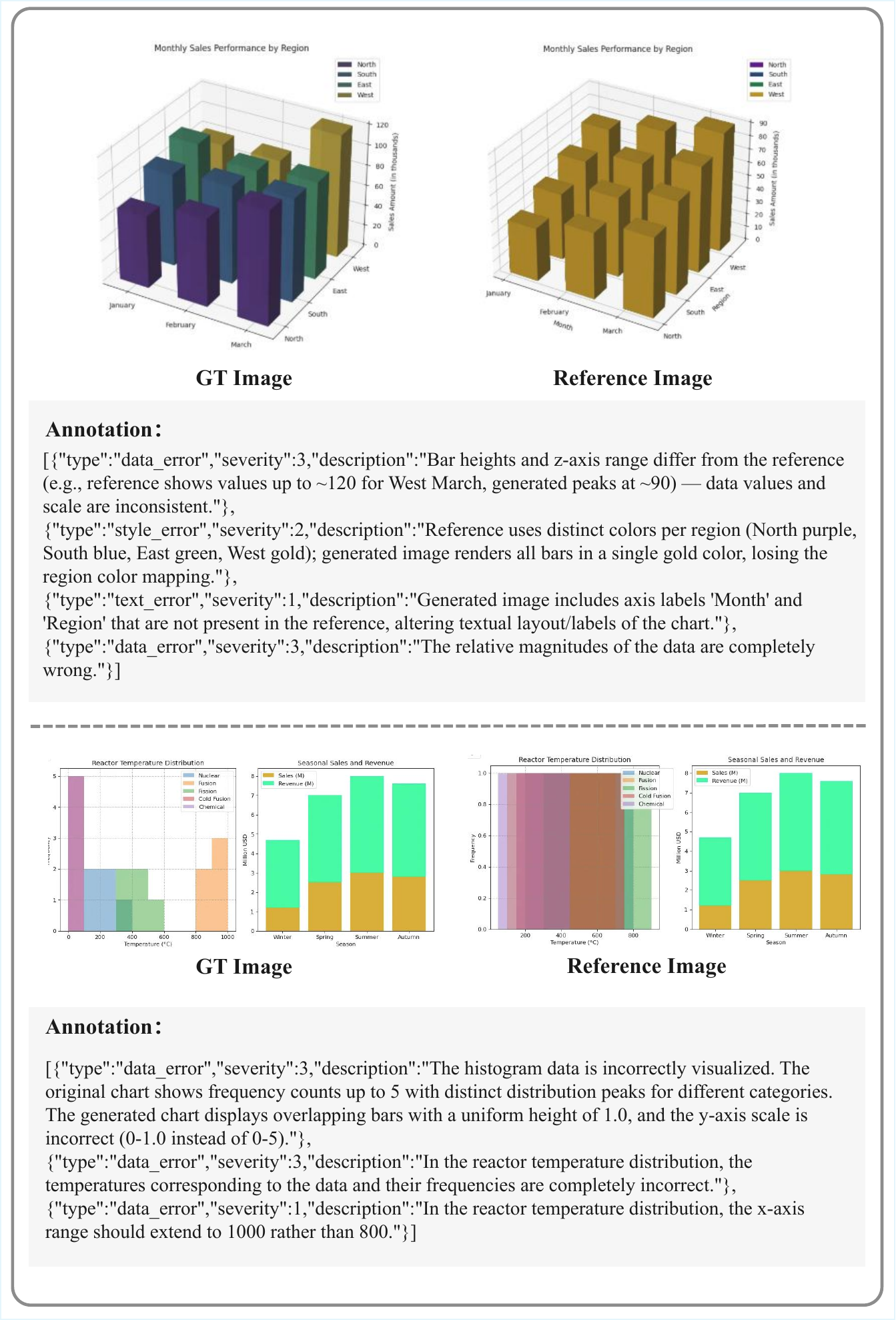}
    \end{center}
    \vspace{-12pt}
    \caption{\small \textbf{Case study.} We show representative examples from the Chart-to-Code reward-model training data and the benchmark.}
    \label{fig:Sup_chart_case}
\end{figure*}

\begin{figure*}[http]
    \begin{center}
    \includegraphics[width=1.\linewidth]{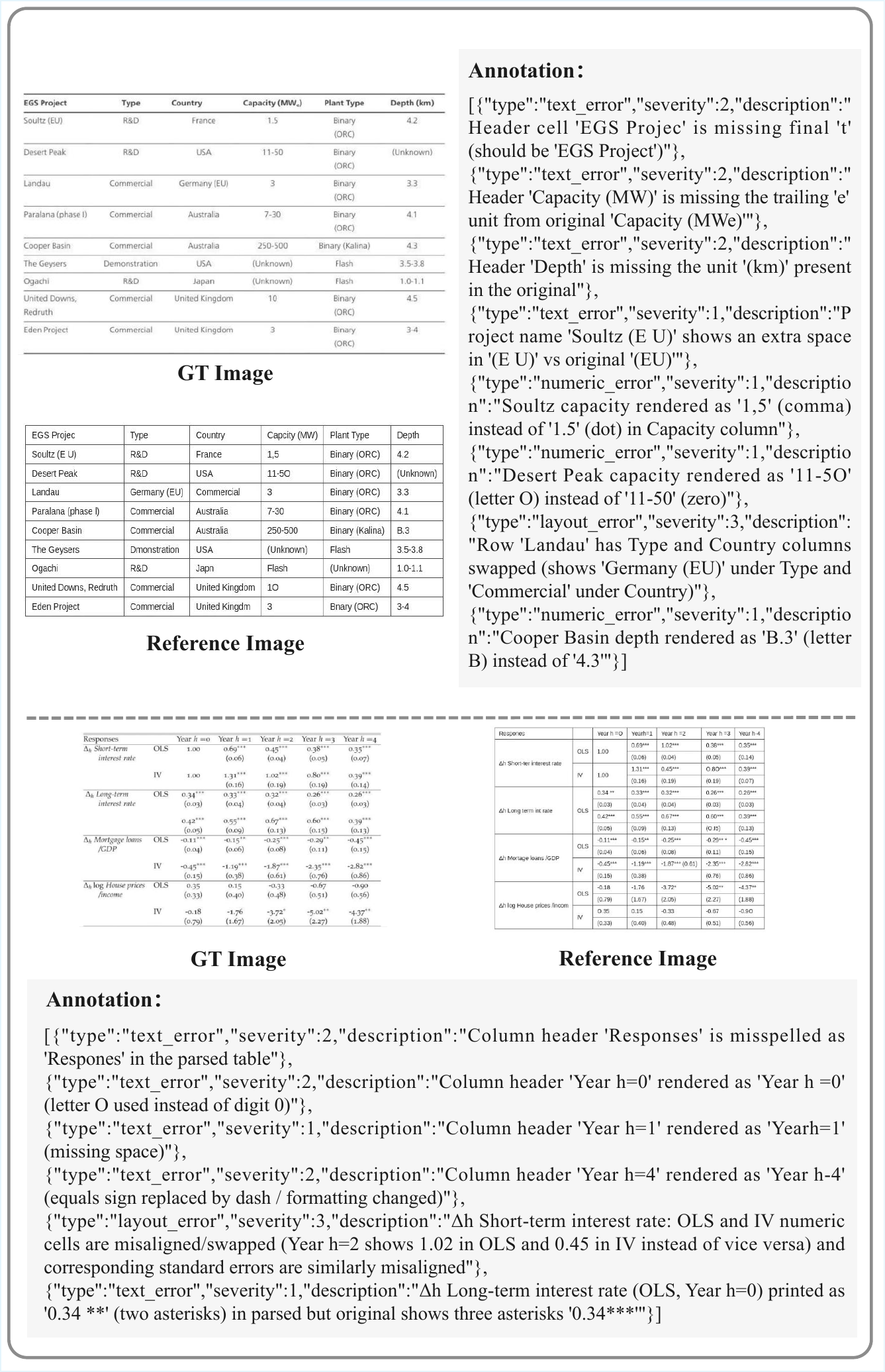}
    \end{center}
    \vspace{-12pt}
    \caption{\small \textbf{Case study.} We show representative examples from the Table-to-Markdown reward-model training data and the benchmark.}
    \label{fig:Sup_table_case}
\end{figure*}

\begin{figure*}[http]
    \begin{center}
    \includegraphics[width=1.\linewidth]{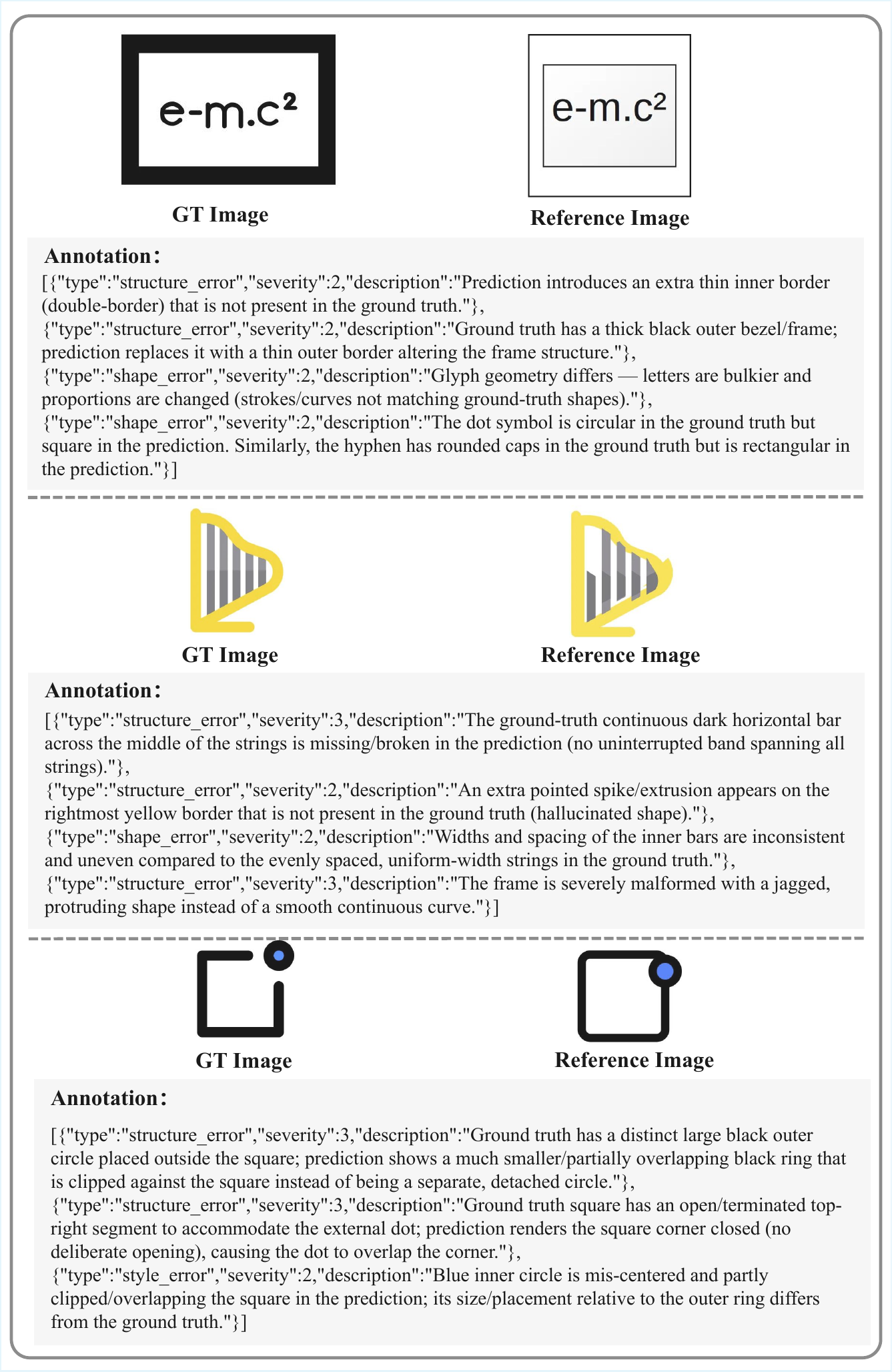}
    \end{center}
    \vspace{-12pt}
    \caption{\small \textbf{Case study.} We show representative examples from the SVG-to-Code reward-model training data and the benchmark.}
    \label{fig:Sup_svg_case}
\end{figure*}

\begin{figure}[t]
    \centering
    \includegraphics[width=\linewidth]{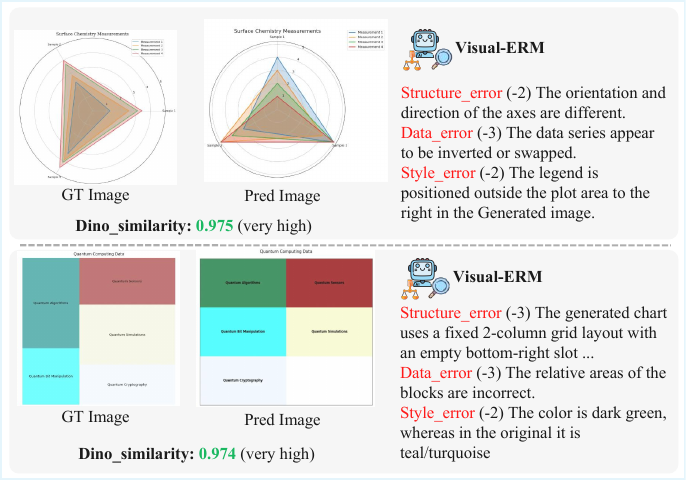}
    \vspace{-12pt}
    \caption{\small \textbf{Further Case Analysis.} We present additional cases to highlight the  limitations of current reward models.}
    \vspace{-12pt}
    \label{fig:dino_case}
\end{figure}

\section{Case Study}\label{sec:case_study}

\subsection{Data Cases}
\label{sec:appendix_case}

In this section, we provide representative qualitative examples to illustrate the data format and error annotations used in our pipeline. The reward-model training data and \benchname examples follow a similar structure---both are built from (Original Image, Re-rendered Image) pairs and accompanied by fine-grained error descriptions. Compared to the training data, \benchname is labeled by stronger proprietary models and further curated through human filtering, making it more reliable for evaluation. We showcase examples from \benchname across three domains (Chart, Table, and SVG), as shown in Fig.~\ref{fig:Sup_chart_case}, Fig.~\ref{fig:Sup_table_case}, and Fig.~\ref{fig:Sup_svg_case}.

\paragraph{Chart cases (Fig.~\ref{fig:Sup_chart_case}).}
The chart examples highlight typical failure modes where likely textual and structural text are insufficient, and the discrepancies must be judged directly in the rendered space.
In the first example, the annotations capture multiple error types: a high-severity \texttt{data\_error} indicating mismatched bar magnitudes and/or incorrect $z$-axis scaling, a \texttt{style\_error} where the color mapping collapses into a single palette (losing per-category distinction), and a \texttt{text\_error} that flags unexpected or misplaced labels (e.g., additional axis/title tokens that alter readability). This case demonstrates that visually faithful reconstruction requires jointly preserving data geometry, semantic encodings (color-to-category), and text placement.
In the second example, the dominant issues are again \texttt{data\_error}: the predicted histogram overlays bars incorrectly and uses an inconsistent axis range, which changes the implied distribution. The annotations emphasize that even when the global chart type appears correct, local numeric scaling and bin/value placement errors can lead to critical semantic distortion.

\paragraph{Table cases (Fig.~\ref{fig:Sup_table_case}).}
The table examples illustrate that errors often concentrate on OCR- and layout-related details, where small differences translate into large structural mistakes in the parsed table.
In the first example, the annotations include multiple \texttt{text\_error} instances on header strings (missing characters, missing units) as well as \texttt{numeric\_error} cases where punctuation or symbols change the value semantics (e.g., comma vs.\ dot in decimals, letter \texttt{O} vs.\ digit \texttt{0}). Additionally, a high-severity \texttt{layout\_error} captures column swapping, which is particularly harmful because it can propagate to many cell assignments and invalidate the table schema.
In the second example, the errors focus on fine-grained text rendering and alignment: header tokens are misspelled, spacing/formatting changes introduce mismatched column names, and significance markers are rendered inconsistently. The \texttt{layout\_error} further reflects misalignment of numeric entries across columns, a common structured-visual failure that is hard to detect in the text space but clearly visible after rendering.

\paragraph{SVG cases (Fig.~\ref{fig:Sup_svg_case}).}
The SVG examples emphasize geometry- and structure-centric discrepancies that arise in vector-graphic reconstruction.
In the first case, the annotations include \texttt{structure\_error} due to an extra inner border and an altered frame composition, along with \texttt{shape\_error} describing changed glyph geometry (stroke thickness and curve profiles) and subtle symbol differences (e.g., rounded vs.\ rectangular caps). 
In the second case, the prediction introduces hallucinated protrusions and fails to reproduce continuous elements (e.g., a missing uninterrupted horizontal band), while also exhibiting inconsistent spacing of repeated bars---a combination of \texttt{structure\_error} and \texttt{shape\_error}.
In the third case, the mismatches include incorrect global composition (outer ring size/placement), an incorrect top-right termination detail, and a mis-centered inner circle, reflecting intertwined \texttt{structure\_error} and \texttt{style\_error}. These cases show that SVG fidelity depends critically on precise spatial relationships, path geometry, and layering, which motivates an image-to-image judge that can reason over fine-grained rendered differences.

Overall, these examples demonstrate that \methodname and \benchname capture diverse, realistic discrepancy patterns across domains, and that the error annotations provide interpretable and actionable supervision signals for both training and evaluation.

\subsection{Further Case Analysis}

We provide additional cases to demonstrate the limitations of existing reward models, as illustrated in Fig.~\ref{fig:dino_case}.



\end{document}